\documentclass[11pt,a4paper]{article}
\usepackage[hyperref]{emnlp2020}
\usepackage{times}
\usepackage{latexsym}
\usepackage{amssymb}		%
\usepackage{graphicx}		%
\usepackage{color}
\usepackage{natbib}
\usepackage{bookmark}
\usepackage{eucal}
\usepackage{amsmath}
\usepackage{multirow}
\usepackage{booktabs}
\usepackage{subcaption}
\usepackage{enumitem}
\usepackage{xspace}
\usepackage{ragged2e}
\usepackage{array}
\def\aclpaperid{1713}
\aclfinalcopy

\newif\ifhidecomments
\hidecommentsfalse

\ifhidecomments
    \newcommand{\joe}[1]{}
    \newcommand{\chenhao}[1]{}
\else
    \newcommand{\joe}[1]{{\color{red}{ #1 --Joe}}}
    \newcommand{\chenhao}[1]{{\color{blue}{\tt #1 --CT}}}
\fi

\newcommand{\para}[1]{\smallskip\noindent{\bf #1}}
\newcommand{\secref}[1]{\S\ref{#1}}
\newcommand{\figref}[1]{Fig.~\ref{#1}}
\newcommand{\tableref}[1]{Table~\ref{#1}}

\begin{document}
\title{Characterizing the Value of Information in Medical Notes}
\author{
    Chao-Chun Hsu$^1$,~~Shantanu Karnwal$^2$,\\ 
    \textbf{Sendhil Mullainathan$^3$,~~Ziad Obermeyer$^4$,~~Chenhao Tan$^2$} \\ 
    $^1$ University of Chicago,$^2$ University of Colorado Boulder \\ 
    $^3$ Chicago Booth School of Business,~~$^4$ University of California, Berkeley \\
    \texttt{chaochunh@uchicago.edu}\\ \texttt{\{shantanu.karnwal,chenhao.tan\}@colorado.edu} \\
    \texttt{sendhil@chicagobooth.edu,~zobermeyer@berkeley.edu}
        }
\date{December 2019}

\maketitle
\begin{abstract}
Machine learning models depend on the quality of input data.
As electronic health records are widely adopted,
the amount of data in health care is growing, along with complaints about the quality of medical notes.
We use two prediction tasks, readmission prediction and in-hospital mortality prediction, to characterize the value of information in medical notes.
We show that as a whole, medical notes only provide additional predictive power over structured information in readmission prediction.
We further propose a probing framework to select parts of notes that enable more accurate predictions than using all notes, 
despite that the selected information leads to a distribution shift from the training data (``all notes'').
Finally, we demonstrate that models trained on the selected {\em valuable} information achieve even better predictive performance, with only 6.8\% 
of all the tokens for readmission prediction.
\end{abstract}

\section{Introduction}

As electronic health records (EHRs) are widely adopted in health care, medicine is increasingly an information science \citep{stead2011biomedical,shortliffe2010biomedical,krumholz2014big}:
{\em Obtaining} and {\em analyzing} information
is critical for the diagnosis, prognosis, treatment, and prevention of disease.
Although EHRs may increase the accuracy of storing structured information (e.g., lab results),
there are growing {\em complaints} about unstructured medical notes (henceforth ``{\em notes}'')
\citep{gawande_2018,payne2015report,hartzband2008off}.

These complaints can be grouped into two perspectives: consumption and production.
On the one hand, information overload poses a critical challenge on the consumption side.
That is,
the sheer amount of information makes it difficult to glean meaningful information from EHRs, including notes \citep{weir2007critical}.

On the other hand, from the perspective of production, for every hour spent on patient interaction, physicians have an added one-to-two hours finishing the progress notes and reviewing results among other things, without extra compensation \citep{patel2018factors}.
The additional work contributes to physician burnout, along with low-quality notes and even errors in the notes.
Consequently, physicians tend to directly copy large volumes of patient data into notes, but may fail to record information {\em only} available through interaction with patients.
For instance, they may miss the wheezing breath for the diagnosis of the chronic obstructive pulmonary disease, or fail to have engaging conversations for evaluating signs of depression \citep{zeng2016opinion}.

While the NLP community has focused on alleviating the challenges in analyzing information (e.g., information overload),
we argue that it is equally important to help caregivers {\em obtain} and {\em record} valuable information in the first place.
We aim to take a first step towards this direction by characterizing the value of information in medical notes computationally.
In this work, we define {\em valuable information} as information that is useful for evaluating medical conditions and making medical decisions.

To do that, we first examine the value of 
notes {\em as a whole} conditioned on structured information.
While narrative texts can {\em potentially} provide valuable information only accessible through physician-patient interaction, 
our analysis addresses the typical complaint that notes contain too many direct copies of structured information such as lab results.
Therefore, a natural question is whether notes provide additional predictive power for 
medical decisions beyond structured information.
By systematically studying two critical tasks, readmission prediction and 
in-hospital mortality prediction, we demonstrate that 
notes are valuable for readmission predictions, but not useful for mortality prediction.
Our results differ from previous studies demonstrating the effectiveness of notes in mortality prediction, partly because \citet{ghassemi2014unfolding} %
use a limited set of structured variables and thus achieve limited predictive power with structured information alone.

We then develop a probing framework to evaluate the prediction performance of {\em parts} of notes selected by value functions.
We hypothesize that 
not all components of notes are equally valuable
and 
some parts of notes can provide stronger predictive power than the whole.
We find that discharge summaries are especially predictive for readmission, while 
nursing notes are most valuable for mortality prediction.
Furthermore, we leverage hypotheses from the medical literature to develop interpretable value functions to identify 
valuable sentences in notes.
Similarity with prior notes turns out to be powerful: 
a mix of most and least similar sentences provide better performance than using all notes, despite containing only a fraction of tokens.

Building on these findings, we finally demonstrate the power of valuable information beyond the probing framework.
We show that classification models trained on the selected valuable information alone provide even better predictive power than using all notes.
In other words, our interpretable value functions can effectively filter noisy information in notes and lead to better models.

We hope that our work encourages future work in understanding the value of information and ultimately improving the quality of medical information obtained and recorded by caregivers, because information is after all created by {\em people}.

\section{Our Predictive Framework}

We investigate the value of notes through a predictive framework.
We consider two prediction tasks using MIMIC-III:\footnote{MIMIC official website: \url{https://mimic.physionet.org/}.} readmission prediction and mortality prediction.
For each task, we examine two questions:
1) does a model trained on both notes and structured information outperform the model with structured information alone? (\secref{sec:overall})
2) using a model trained on all notes, are there interpretable ways to identify parts of notes that are more valuable than all notes? (\secref{sec:value})
\subsection{An Overview of MIMIC-III}
\label{sec:mimic}

MIMIC-III is a freely available medical database 
of de-identified patient records.
This dataset includes basic information about patients such as admission details and demographics, which allows us to identify outcomes of interest such as mortality and readmission.
It also contains detailed information that characterizes the patients' health history at the hospital, known as {\em events}, including laboratory events, charting events, and medical notes.
The data derived from these events are elicited while 
patients are in the hospital.
Our 
goal 
is to characterize the value of such elicited information, in particular, notes, through 
predictive experiments.
Next, we break down the information into two categories:
structured vs. unstructured.

\para{Structured information.}
The structured 
information
includes the numeric and categorical results of 
medical measurements and evaluations of patients. 
For example, in MIMIC-III, structured information includes status monitoring, e.g., respiration rate and blood glucose, and fluids that have been administered to or extracted from the patients.

\para{Notes (unstructured texts).}
Caregivers, including nurses and physicians, record information 
based on
their interaction with patients in notes. There are fifteen types of notes in MIMIC-III,
including nursing notes and physician notes.
\tableref{tab:note-type} shows the number of notes in each type and their average length.

Not all admissions have notes from caregivers. 
After 
filtering patients under 18 and other invalid data (see details in the supplementary material),
discharge summary appears in most admissions (96.7\%); however, only 0.1\% of admissions have consult notes. 
The most common types of notes include nursing,\footnote{We merge ``Nursing'' and ``Nursing/other'' in MIMIC-III.} radiology, ECG, and physician.
There is also significant variation in length between different types of notes.
For instance, discharge summary is more than 8 times as long as nursing notes.

\begin{figure}[t]
    \centering
    \includegraphics[width=0.9\linewidth]{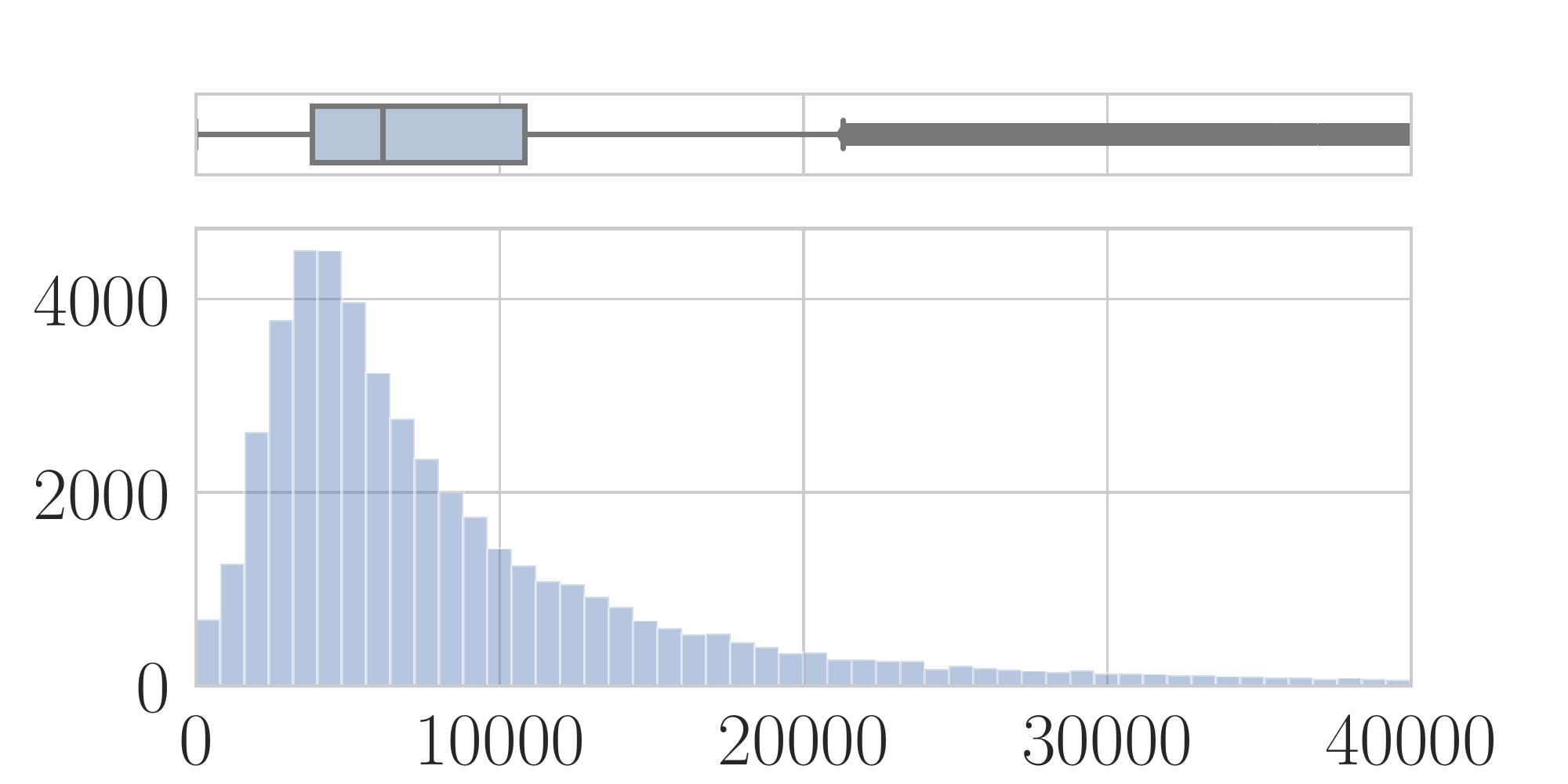}
    \caption{Distribution of token length of admissions. Average: 11,135 tokens. Median: 6,437 tokens.
    }
    \label{fig:length-stats}
\end{figure}

\figref{fig:length-stats} presents the total number of tokens in all types of notes within one admission.
As discussed in the introduction, a huge amount of information (11,135 tokens on average)
is generated in the form of unstructured texts for a patient in an admission.
We hypothesize that not all of them are useful for medical purposes.

\begin{table}[]
	\centering
	\scriptsize
	\resizebox{.9\linewidth}{!}{
	\begin{tabular}{lrrr}
	\toprule
	CATEGORY          & COUNT  & \%   & LEN. \\ \midrule
	Nursing           & 506,528  & 73.0 & 241       \\ 
	Radiology         & 338,834 & 83.3 & 449       \\ 
	ECG               & 123,042 & 61.3 & 43        \\ 
	Physician         & 92,426  & 18.2 & 1369      \\ 
	Discharge summary & 47,572  & 96.7 & 2195      \\ 
	Echo              & 34,064  & 45.8 & 464       \\ 
	Respiratory       & 32,798  & 8.1  & 205       \\ 
	Nutrition         & 7,971   & 6.4  & 602       \\ 
	General           & 7,710   & 6.4  & 290       \\ 
	Rehab Services    & 5,321   & 4.6  & 622       \\ 
	Social Work       & 2,294   & 2.8  & 446       \\ 
	Case Management   & 939    & 1.3  & 260       \\ 
	Pharmacy          & 97     & 0.1  & 512       \\ 
	Consult           & 78     & 0.1  & 1206      \\ 
	\bottomrule
	\end{tabular}
	}
	\caption{Statistics of note events in MIMIC-III after data preprocessing. \% denotes the proportion of admissions having this type of note. ``LEN.'' means the average length.
	}
	\label{tab:note-type}
	\end{table}

\subsection{Task Formulation \& Data Representation}
\label{sec:method}

We consider the following two 
prediction tasks related to important medical decisions.

\begin{itemize}[itemsep=0pt,leftmargin=*]
    \item {\bf Readmission prediction.} We aim to predict whether a patient will be re-admitted to the hospital in 30 days after being discharged, given the information collected within one admission. 

    \item {\bf In-hospital mortality prediction.} We aim to predict whether a patient dies in the hospital within one admission. 
    Following \citet{ghassemi2014unfolding}, we consider three  time periods: 24 hours, 48 hours, and retrospective.
    The task is most difficult but most useful with only information from the first 24 hours.
    We thus focus on that time period in the main paper (see the supplementary material for 48 hours and retrospective results). 
\end{itemize}

Formally, our data is a collection of time series with labels corresponding to each task, $\mathcal{D} = \{(E_i, y_i)\}_{i=1}^{N}$ where $N$ is the number of admissions (instances). For each collection of time series $E = \{(h_t, \tau_t, x_t)\}_{t=1}^{T}$ of an admission, 
$h_t$ represents the timestamp (e.g., $h_t = 4.5$ means 4.5 hours after admission) and $\tau_t \in \{0, 1\}$ captures the type of an event (0 indicates that the event
contains structured variable and 1 indicates that the event is a note) and $x_t$ stores the value of the corresponding event.
Our goal is to predict label $y \in \{0, 1\}$: in readmission, $y$ represents whether a patient was re-admitted within a month.
In mortality prediction, $y$ represents whether a patient died in this admission.\footnote{Discharge summaries are not used for mortality prediction since they may leak the label.}

As a result, we obtained a total of 37,798/33,930 unique patients and 46,968/42,271 admissions for readmission/mortality prediction (24 hours).
\para{Representing structured information.}
As structured information is sparse over timestamps, 
we filter event types that occur less than 100,000 times (767 event types remaining).\footnote{We exclude structured information such as ICD-9 codes that might have been filled in retrospectively.}
Following \citet{harutyunyan2017multitask},
we represent the time series data of structured variables into a vector
by extracting basic statistics of different time windows.
Specifically, for events of structured variables, \mbox{\small $E_i^{\tau=0} = \{(h_t, \tau_t, x_t) | \tau_t=0\}_{t=1}^{T}$} 
where $x_t \in R^{d}, d=767$,
we apply six statistical functions on seven sub-periods 
to generate $e_{i} \in \mathbb{R}^{d\times 7 \times 6}$ as the representation of structured variables. The six statistical functions are maximum, minimum,
mean, standard deviation, skew, and number of measurements. The seven sub-periods are the entire time period, first (10\%, 25\%, 50\%) of the time period, and last (10\%, 25\%, 
50\%) of the time period.
We then impute missing values with the mean of training data
and apply min-max 
normalization. 

\para{Representing notes.}
For notes in an admission, 
we apply sentence and word tokenizers in the NLTK toolkit to each note~\cite{Loper02nltk}.
See \secref{sec:setup} for details on how we use tokenized outcomes for different machine learning models.

\subsection{Experimental Setup}
\label{sec:setup}

Finally, we discuss the experimental setup and models that we explore in this work. 
Our code is available at \url{https://github.com/BoulderDS/value-of-medical-notes}.

\para{Data split.}
Following the training and test split\footnote{Data split can be found here: \url{https://github.com/YerevaNN/mimic3-benchmarks}.} of patients in \citet{Harutyunyan_2019},
we 
use 85\% of the patients for training 
and the rest 15\% for testing.
To generate the validation set, we first split 20\% of the patients from training set 
and then collect the admissions under each patient to prevent information leaking for the same patient. 

\para{Models.}
We consider the following models.

\begin{itemize}[itemsep=0pt,leftmargin=*,topsep=-2pt]
    \item Logistic regression (LR). For notes, we use 
    tf-idf representations. 
    We simply concatenate structured variables with $\ell_2$-normalized tf-idf vector from notes to incorporate structured information.
    We use scikit-learn \cite{scikit-learn} and apply $\ell_2$ regularization to prevent overfitting.
    We search hyperparameters $C$ 
    in \mbox{\small $\{2^x | x \in \mathbb{Z}, -11 \le x \le 0 \}$}.
    \item Deep averaging networks (DAN) \cite{iyyer2015deep}.
    We use the average embedding of all tokens in the notes to represent the unstructured information, which can be considered a deep version of bag-of-words methods.
    Similar to logistic regression, we concatenate the structured variables with the average embedding of words in notes to incorporate structured information.
    \item GRU-D \cite{che2018recurrent}.
    The key innovation of GRU-D is to account for 
    missing data in EHRs.
    It imputes the missing value by considering all the information available so far, including how much time it elapses since the last observation and all the previous history.
    Similar to DAN, we use the average embedding of tokens to represent notes.
    See details of 
    GRU-D 
    in the supplementary material.\\
\end{itemize}

Although it is difficult to apply the family of BERT models to this dataset due to 
their input length limitation compared to the
large number of tokens 
from all medical notes,
we experiment with ClinicalBERT \cite{alsentzer-etal-2019-publicly} based on the selected valuable information 
in \secref{sec:value}.

\para{Evaluation Metrics}
ROC-AUC is often used in prior work on MIMIC-III \citep{ghassemi2014unfolding,harutyunyan2017multitask}.
However, when the number of negative instances is much larger than positive instances, the false positive rate in ROC-AUC becomes insensitive 
to the change of false positive instances.
Therefore, area under precision-recall curve (PR-AUC) 
is considered
more informative than ROC-AUC~\cite{davis2006relationship},
In our experiments, the positive fraction is only 7\% and 12\% in readmission prediction and mortality prediction respectively.\footnote{To emulate class ratios in real data, we do not subsample to achieve balanced data.}
As precision 
is often critical in medical decisions, we also present precision at 1\% and 5\%.

\section{Do Medical Notes Add Value {\em over} Structured Information?}
\label{sec:overall}

\begin{figure*}[!t]
    \centering
    \large{Readmission Prediction}\\
    \vspace{0.7em}
    \begin{subfigure}[t]{0.18\textwidth}
        \centering
        \includegraphics[trim=0 0 1020 0,clip,width=0.99\textwidth]{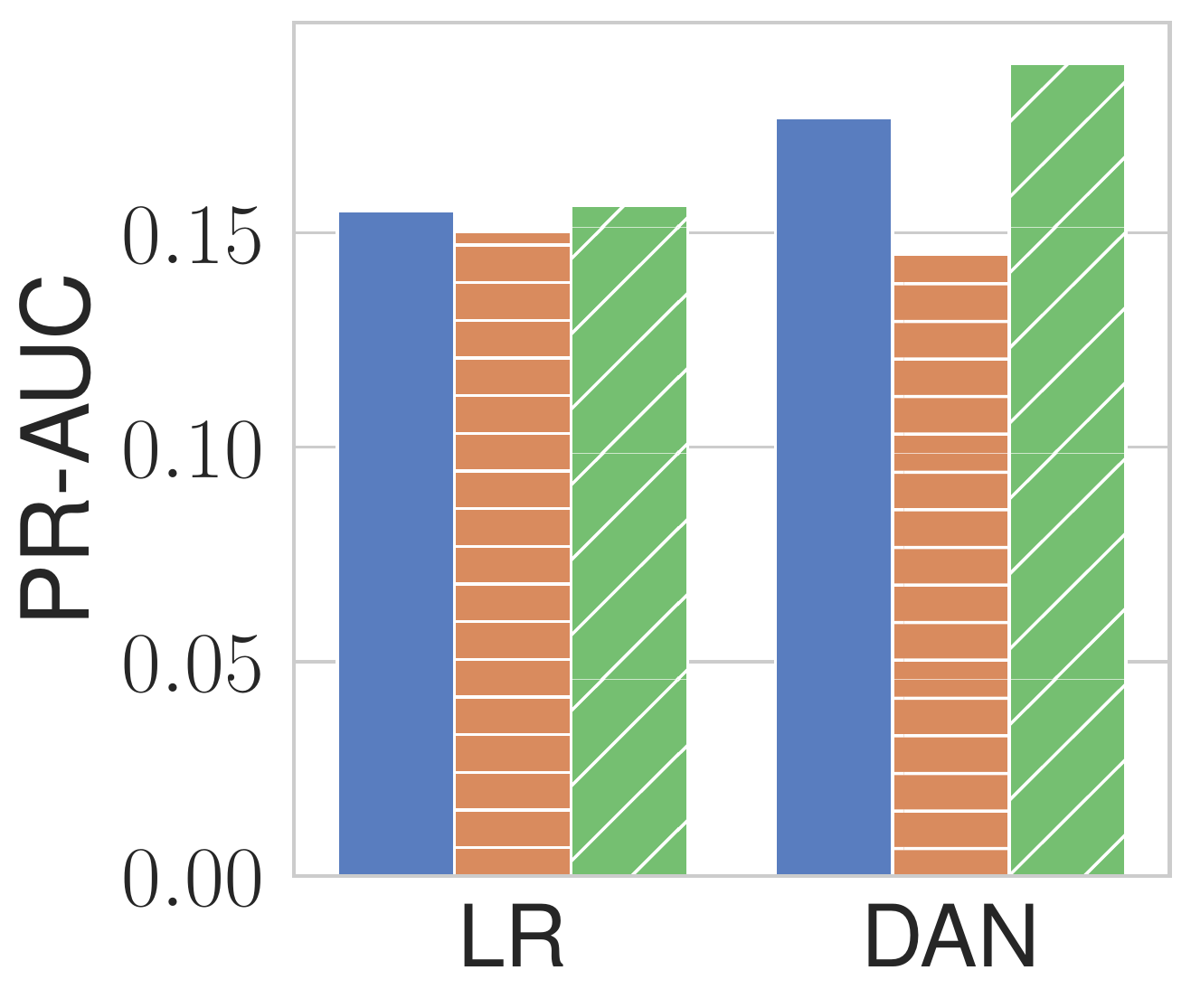}
        \caption{PR-AUC.}
        \label{fig:prauc-readmission}
    \end{subfigure}
    \begin{subfigure}[t]{0.18\textwidth} 
        \centering
        \includegraphics[trim=0 0 1020 0,clip,width=0.95\textwidth]{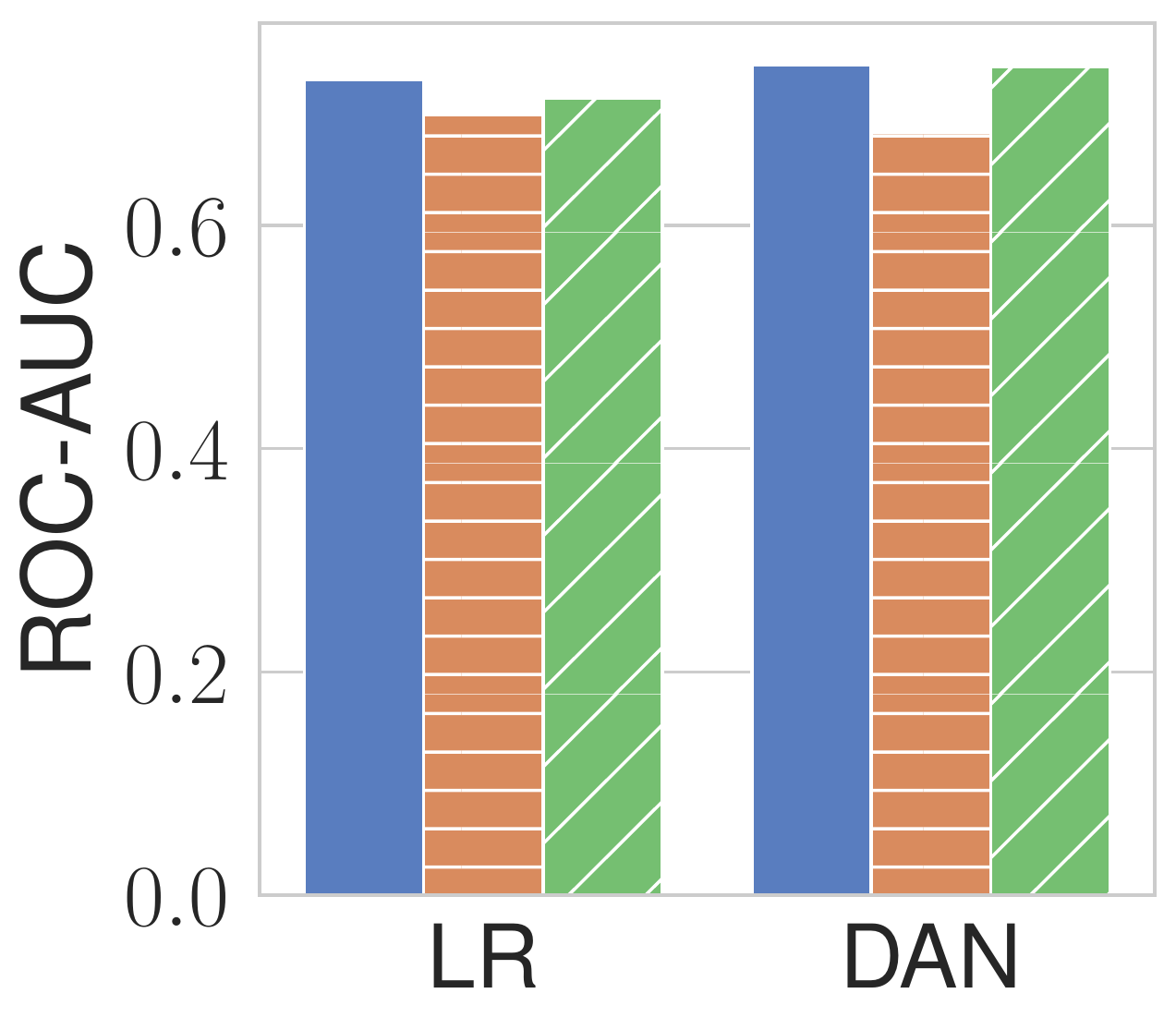}
        \caption{ROC-AUC.}
        \label{fig:rocauc-readmission}
    \end{subfigure}
    \begin{subfigure}[t]{0.18\textwidth}
        \centering
        \includegraphics[trim=0 0 1020 0,clip,width=0.95\textwidth]{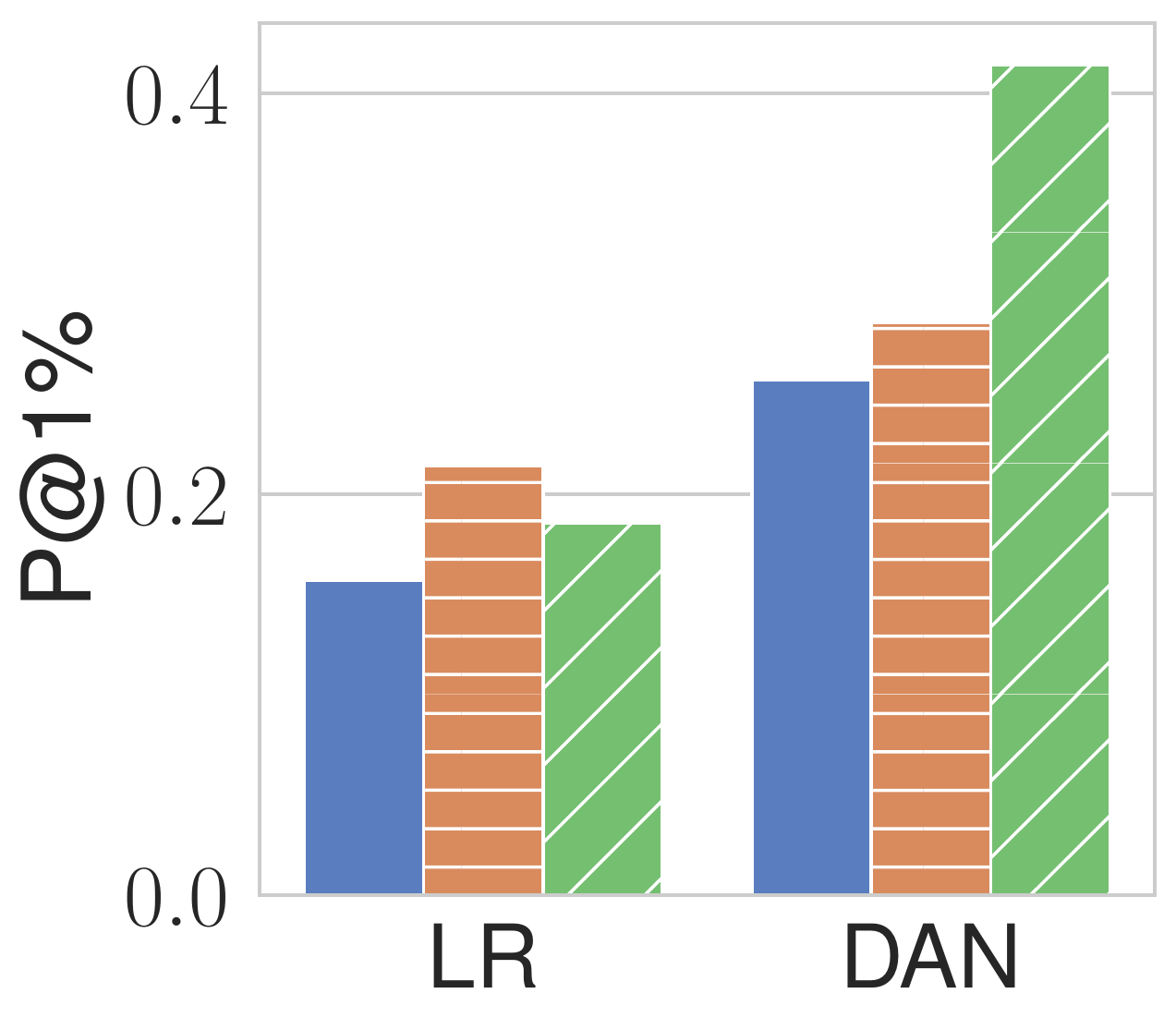}
        \caption{Precision at 1\%.}
        \label{fig:p1-readmission}
    \end{subfigure}
    \begin{subfigure}[t]{0.18\textwidth}
        \centering
        \includegraphics[trim=0 0 1020 0,clip,width=0.95\textwidth]{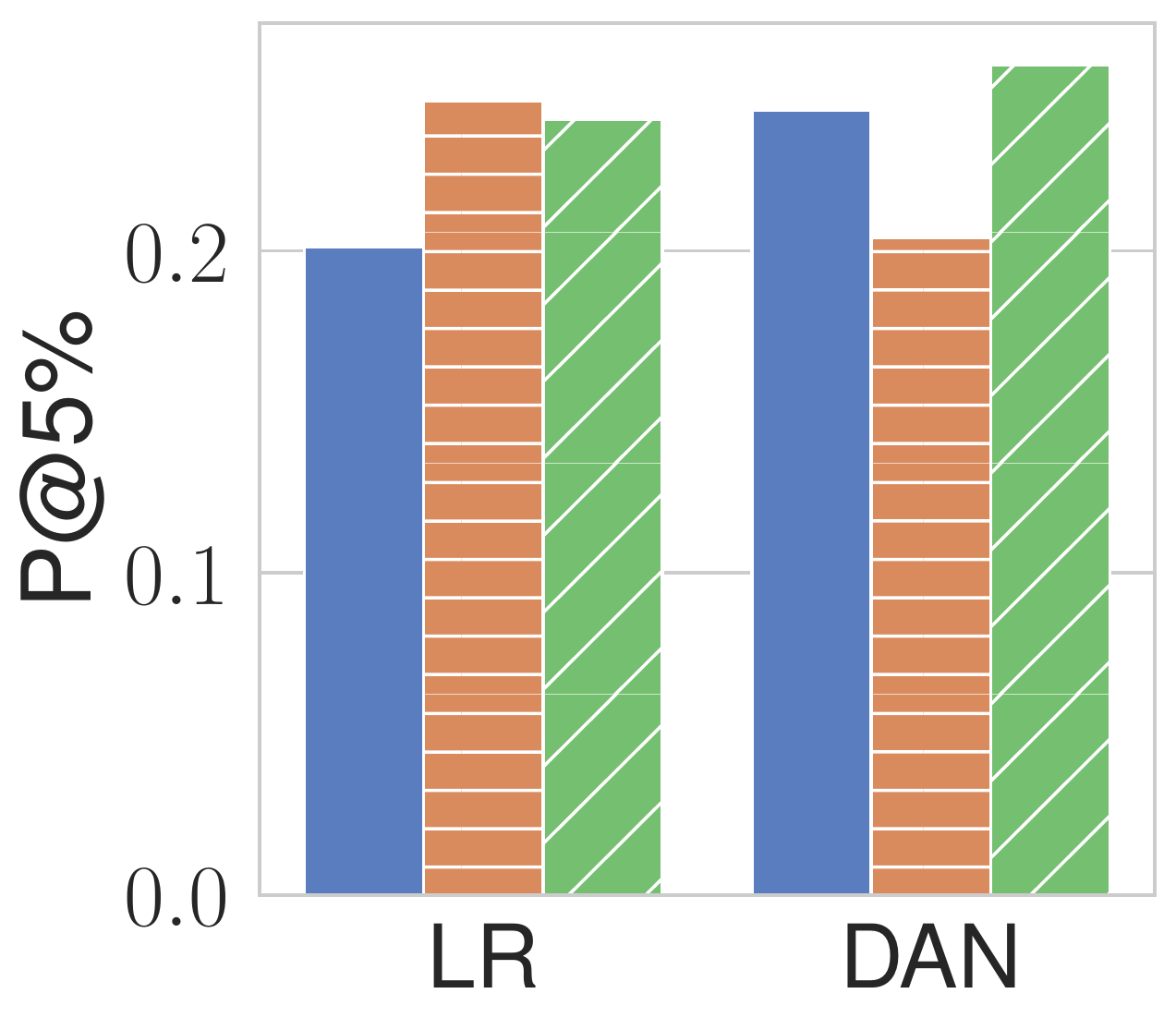}
        \caption{Precision at 5\%.}
        \label{fig:p5-readmission}
        \vspace{0.7em}
    \end{subfigure}

    \large{In-hospital Mortality Prediction (24 hours)}\\
    \begin{subfigure}[t]{0.18\textwidth}
        \centering
        \includegraphics[trim=0 0 1020 0,clip,width=0.95\textwidth]{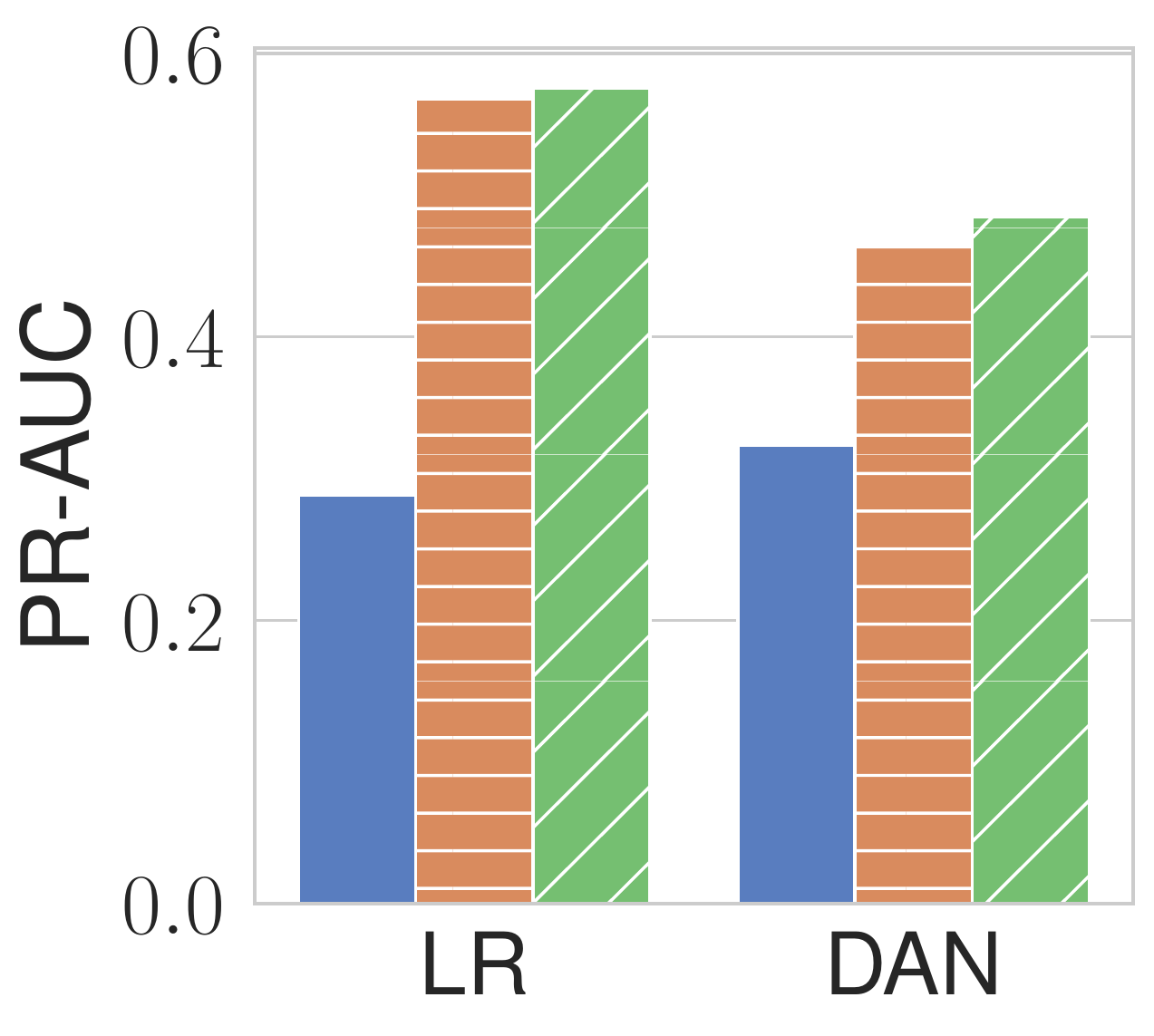}
        \caption{PR-AUC.}
        \label{fig:prauc-mortality-main}
    \end{subfigure}
    \begin{subfigure}[t]{0.18\textwidth}
        \centering
        \includegraphics[trim=0 0 1020 0,clip,width=0.99\textwidth]{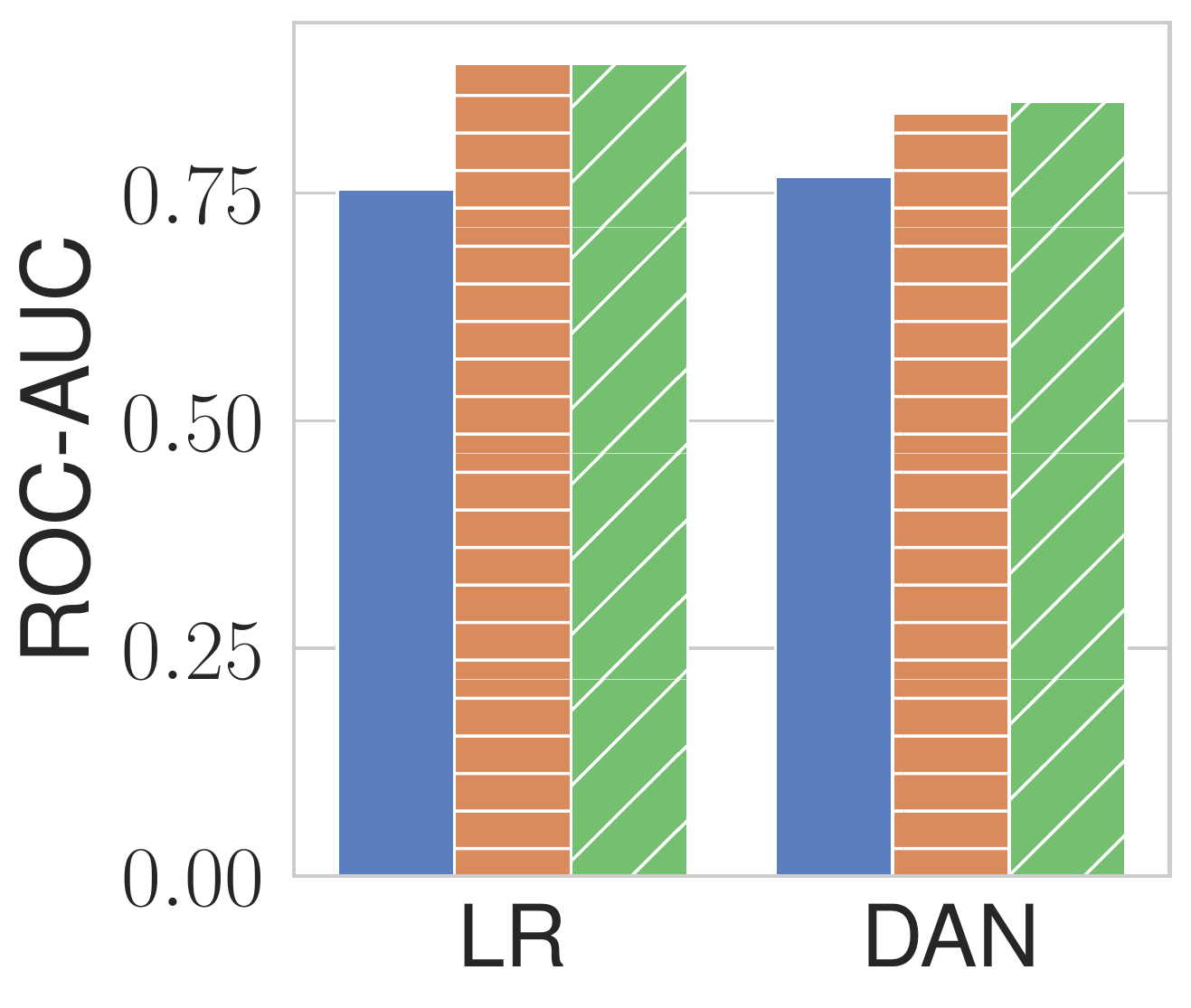}
        \caption{ROC-AUC.}
        \label{fig:rocauc-mortality-main}
    \end{subfigure}
    \begin{subfigure}[t]{0.18\textwidth}
        \centering
        \includegraphics[trim=0 0 1020 0,clip,width=0.99\textwidth]{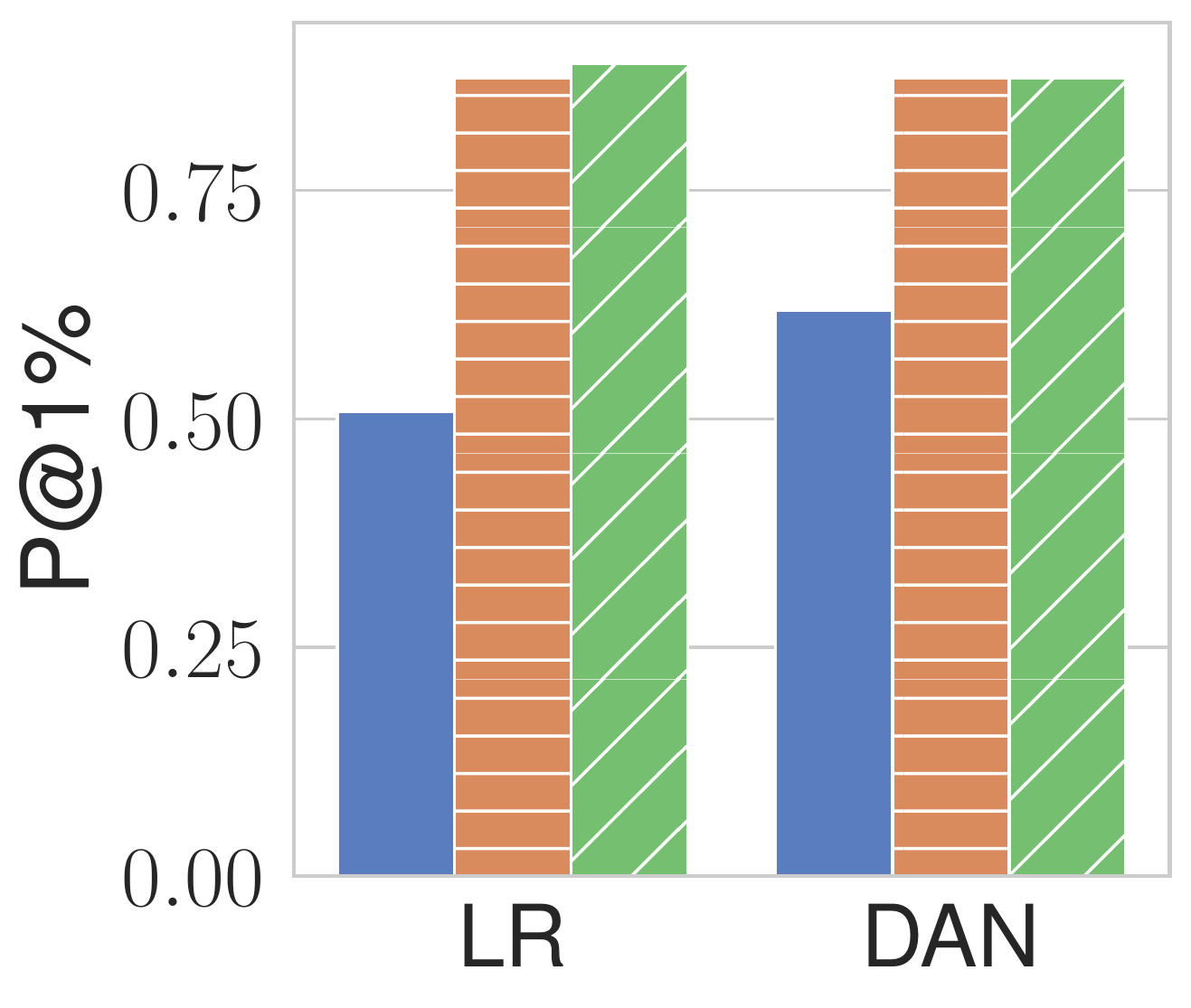}
        \caption{Precision at 1\%.}
        \label{fig:p1-mortality-main}
    \end{subfigure}
    \begin{subfigure}[t]{0.18\textwidth}
        \centering
        \includegraphics[trim=0 0 1020 0,clip,width=0.95\textwidth]{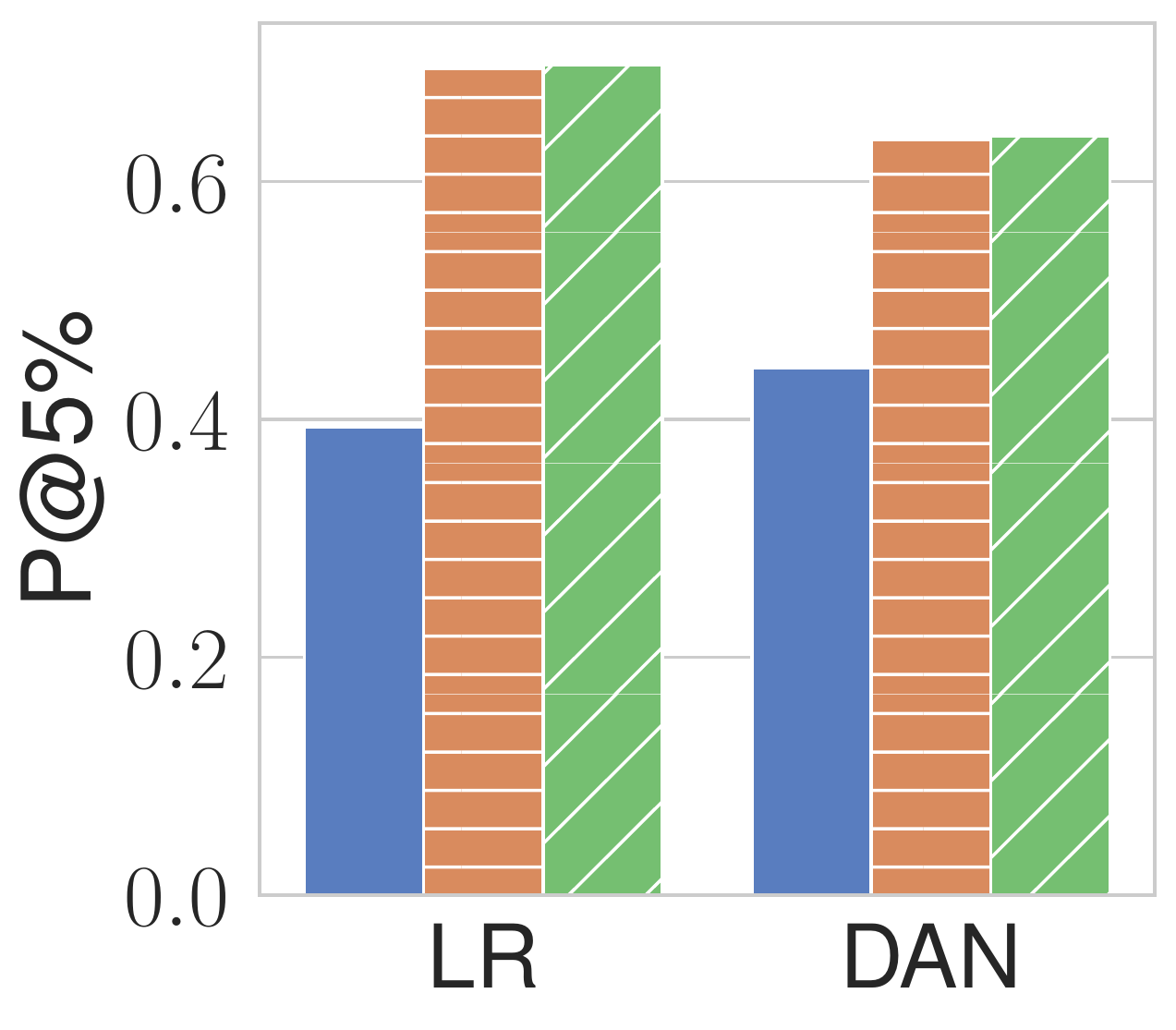}
        \caption{Precision at 5\%.}
        \label{fig:p5-mortality-main}
    \end{subfigure}
    \begin{subfigure}[t]{.9\textwidth}
        \centering
        \includegraphics[trim=365 195 0 40,clip,width=0.7\textwidth]{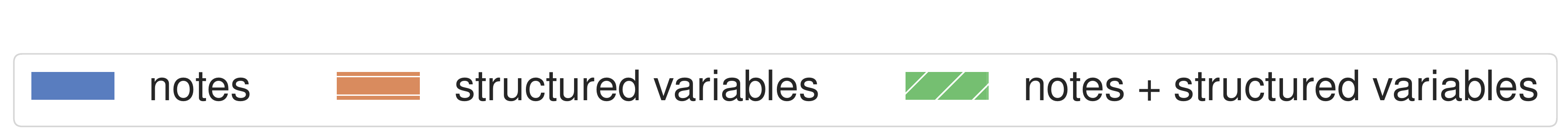}
    \end{subfigure}
    \caption{Results of PR-AUC/ROC-AUC/Precision at 1\%/Precision at 5\% on logistic regression (LR)/deep averaging networks (DAN) models in readmission prediction and mortality prediction (24 hours). 
    Notes are valuable for readmission predictions, but are marginally valuable in mortality prediction.
    }
    \label{fig:results-main}
\end{figure*}

Our first 
question is concerned with whether medical notes provide any additional predictive value over structured variables.
To properly address this question, we need a strong baseline with structured information.
Therefore, we include 767 
types of structured variables to represent structured information (\secref{sec:method}).
Overall, our results are mixed for readmission prediction and in-hospital mortality prediction.
We present results from GRU-D in the supplementary material because GRU-D results reveal similar trends and usually underperform logistic regression or DAN in our experiments.

\para{Notes outperform structured variables in PR-AUC and ROC-AUC  in readmission prediction (\figref{fig:prauc-readmission}-\ref{fig:p5-readmission}).}
For both logistic regression and DAN, notes are more predictive than structured information in readmission prediction based on PR-AUC and ROC-AUC.
In fact, in most cases, structured variables provide little additional predictive power over notes (except PR-AUC with DAN).
Interestingly, we observe mixed results for precision-based metrics.
Structured information can outperform notes in identifying the patients that are most likely to be readmitted.
For DAN, combining notes and structured information provides a significant boost in precision at 1\% compared to one type of information alone, with an improvement of 16\% and 13\% in absolute precision over notes and structured variables respectively.

\para{Structured information dominates notes in mortality prediction (\figref{fig:prauc-mortality-main}-\ref{fig:p5-mortality-main}).}
We 
observe marginally additional predictive value in mortality prediction by incorporating notes with structured information.
In our experiments, the improvement is negligible across all metrics.
This result differs from \citet{ghassemi2014unfolding}.
We believe that the reason is that \citet{ghassemi2014unfolding} only consider age, gender, and the SAPS II score in structured information, while our work considers substantially more structured variables.
It is worth noting that logistic regression with our complete set of structured variables provides better performance than DAN and the absolute number in ROC (0.892) 
is 
better than the best number (0.79) 
in prior work \cite{che2018recurrent}.
The reason for the limited value of notes might be that mortality prediction is a relatively simple task where structured information provides unambiguous signals.

In sum, we find that note 
contributes valuable information over structured variables in readmission prediction, but almost no additional value in mortality prediction. 
Note that ROC-AUC tends to be insensitive to different models and information.
We thus use PR-AUC in the rest of the work to discuss the value of selected information.

\section{Finding Needles in a Haystack: Probing for the Valuable Information}
\label{sec:value}

The key goal of this work is to identify valuable components within notes, as we hypothesize that not all information in notes is valuable for medical decisions, as measured by the predictive power.

To identify valuable components, we leverage an existing machine learning model (e.g., models in \figref{fig:results-main}) and hypothesize that the test performance is better if we only use the ``valuable'' components.
Formally, assume that we trained a model using all notes, $f_{\text{all}}$. 
$S_i$ denotes sentences in all notes ($E_i^{\tau=1}= \{(h_t, \tau_t, x_t) | \tau_t=1\}_{t=1}^{T}$) for an admission in the test set.
We would like to find a subset of sentences $s_i \subset S_i$ so that $f_{\text{all}}(s_i)$ provides more accurate predictions than $f_{\text{all}}(S_i)$.
Note that $s_i$ by definition entails a distribution shift from the data that $f_{\text{all}}$ is trained on ($S_i$), because $s_i$ is much shorter than $S_i$.

The challenge lies in developing interpretable ways to identify 
valuable content.
We first compare the value of different types of notes in \secref{sec:type_cmp}, which can be seen as trivial value functions based on type of note,
and then propose interpretable value functions to zoom in on the content of notes (\secref{sec:heuristics}).
Finally, we show that these valuable components not only provide accurate predictions with a model trained with all the notes, but also allow us to learn a model with better predictive power than that trained with all the notes (\secref{sec:train_value}).
In other words, we can effectively remove the noise by focusing on the valuable components.

\begin{figure*}[!t]
  \centering
  \begin{subfigure}[t]{0.3\textwidth}
    \centering
    \includegraphics[width=\textwidth]{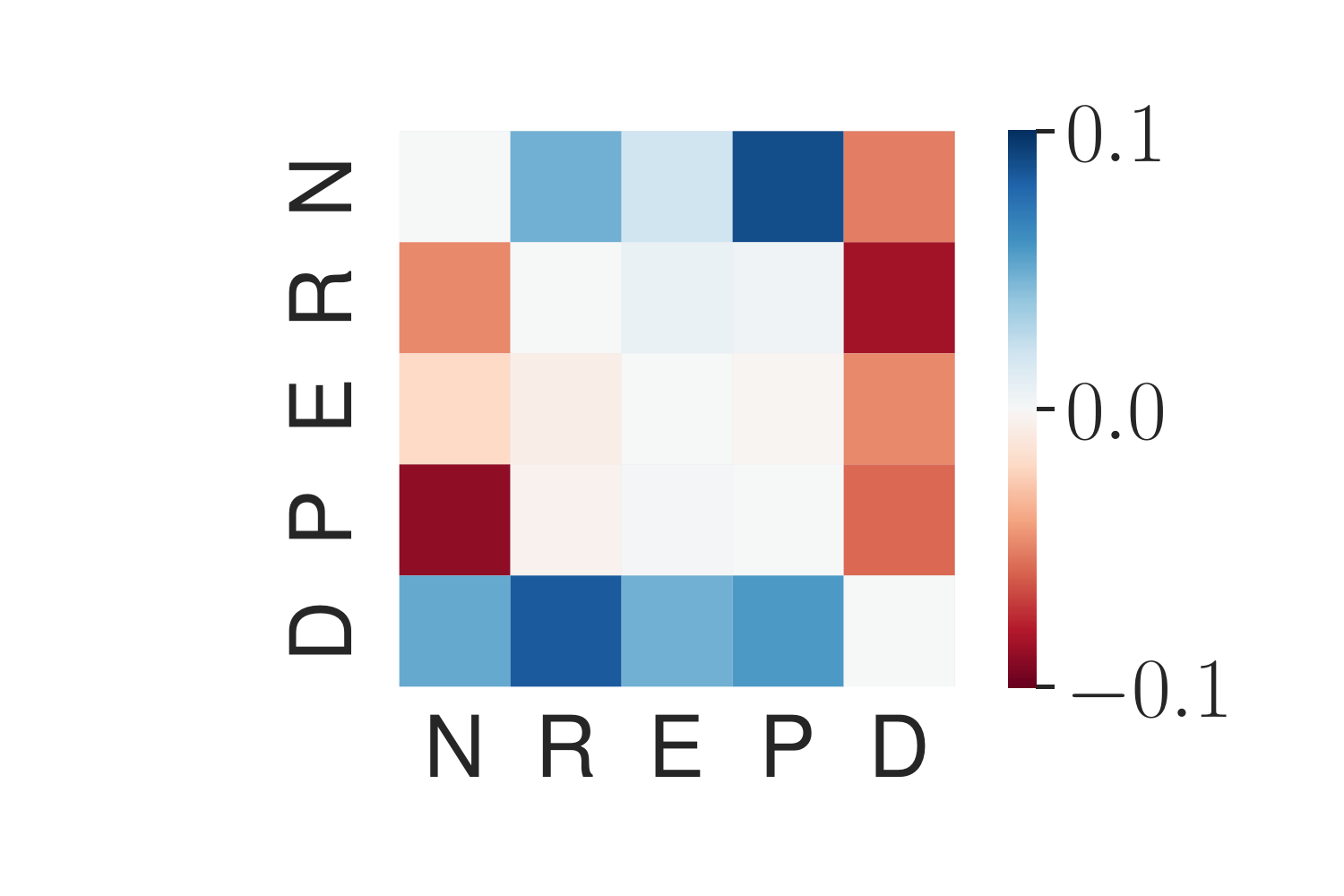} 
    \caption{Readmission prediction.}
    \label{fig:pairwise-readmission-LR}
  \end{subfigure}
  \hfill
  \begin{subfigure}[t]{0.3\textwidth}
    \centering
    \includegraphics[width=\textwidth]{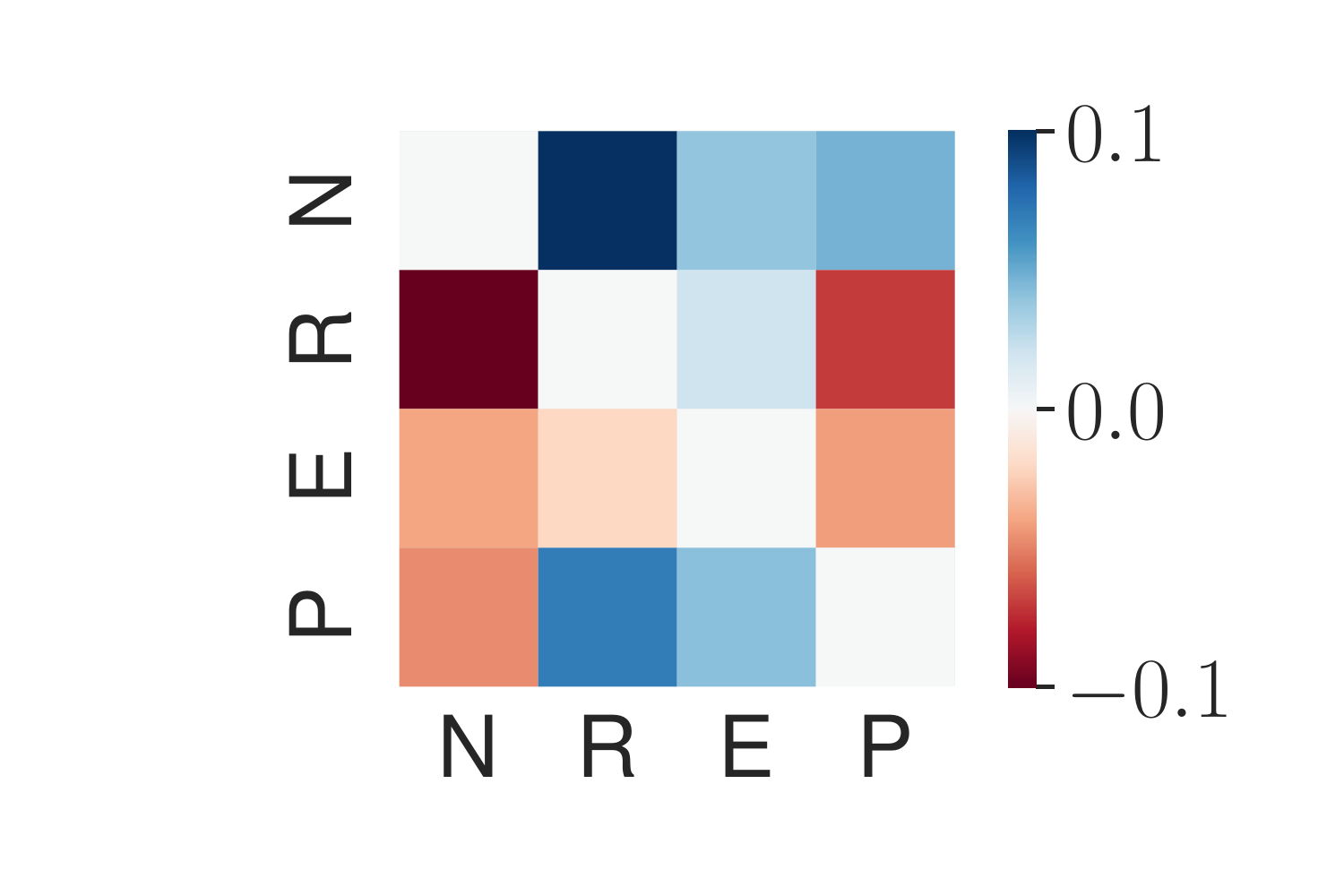}
    \caption{Mortality prediction (24 hrs).}
    \label{fig:pairwise-mortality}
  \end{subfigure}
  \hfill
  \begin{subtable}{0.25\textwidth}
  \vspace{-1in}
  \begin{tabular}{l}
  \toprule
  P: Physician notes \\
  N: Nursing notes \\
  D: Discharge summary\\
  R: Radiology reports \\
  E: ECG reports \\
  \bottomrule
  \end{tabular}
  \end{subtable}
  \caption{Pairwise comparisons between different types of note with logistic regression (each cell shows \mbox{\small $\operatorname{PR-AUC}(f_{\text{all}}(s_{t_{\operatorname{row}}}), y) - \operatorname{PR-AUC}(f_{\text{all}}(s_{t_{\operatorname{column}}}), y))$}.
  To account for the differences in length, we 
  subsample two types of note under comparison to be the same length and report the average values of 10 samples.
  Discharge summaries dominate all other types of notes in readmission prediction, while nursing notes are most useful for mortality prediction.
  }
  \label{fig:pairwise-main}
  \end{figure*}

\subsection{Discharge Summaries, Nursing Notes, and Physician Notes are Valuable}
\label{sec:type_cmp}

To answer our first question, we compare the effectiveness of different types of notes within the top five most common categories:
nursing, radiology, ECG, physician, and discharge summary.
An important challenge lies in the fact that not every admission produces all types of notes.
Therefore, we conduct pairwise comparison that ensures an admission has both types of note.
Specifically, for each pair of note types $(t_1, t_2)$, we choose admissions with both two types of note and make predictions using $s_{t_1}$ and $s_{t_2}$ respectively, where $s_t$ refers to all the sentences in notes of type $t$. 
Each cell in \figref{fig:pairwise-main} indicates \mbox{\small $\operatorname{PR-AUC}(f_{\text{all}}(s_{t_{\operatorname{row}}}), y) - \operatorname{PR-AUC}(f_{\text{all}}(s_{t_{\operatorname{column}}}), y)$} with LR (see the supplementary material for DAN results, which are similar to LR). 
For instance, the top right cell in \figref{fig:pairwise-readmission-LR} shows the performance difference between using only nursing notes and using only discharge summaries for admissions with both nursing notes and discharge summaries.
The negative value suggests that nursing notes provide less 
accurate predictions (hence less valuable information) than discharge summaries in readmission prediction.
Note that due to 
significant variance in length between 
types of note,
we subsample $s_{t_{\operatorname{row}}}$ and $s_{t_{\operatorname{column}}}$ to be the same length in these experiments.

\para{Discharge summaries dominate other types of notes in readmission prediction.}
Visually, most of the dark values in \figref{fig:pairwise-readmission-LR} are associated with discharge summaries.
This makes sense because discharge summaries provide a holistic view of the entire admission and are likely most helpful for predicting future readmission.
Among the other four types of notes, nursing notes are the second most valuable.
In comparison, physician notes, radiology reports, and ECG reports are less valuable.

\para{Nursing notes and physician notes are more valuable for mortality prediction.}
For mortality prediction, nursing notes provide the best predictive power%
.
ECG reports always have the worst results. 
Recall that we subsample each type of notes to the same length.
Hence, the lack of value in ECG reports cannot be attributed to its short length.

In summary, templated notes such as radiology reports and ECG reports are 
less valuable for predictive tasks in medical decisions.
While physician notes are 
the central subject in prior work \citep{weir2007critical}, nursing notes are as important for medical purposes
given that there are many more nursing notes and they record patient information frequently.

\begin{figure*}
  \centering
    \begin{subfigure}[t]{0.45\textwidth}
    \centering
    \includegraphics[trim=0 -10 0 0,clip,width=0.8\textwidth]{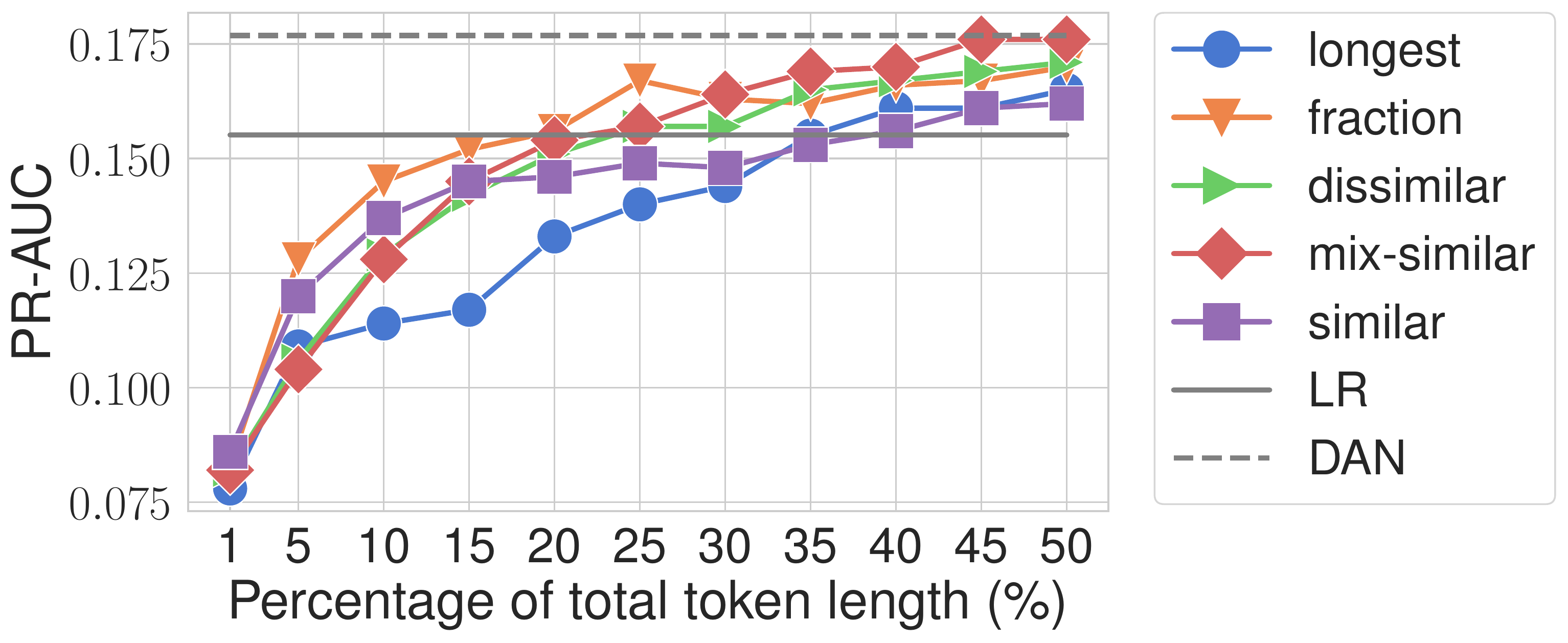}
    \caption{Readmission prediction.}
    \label{fig:heuristics-lr-readmission-main}
  \end{subfigure}
  \begin{subfigure}[t]{0.45\textwidth}
    \centering
    \includegraphics[trim=0 0 0 0,clip,width=0.8\textwidth]{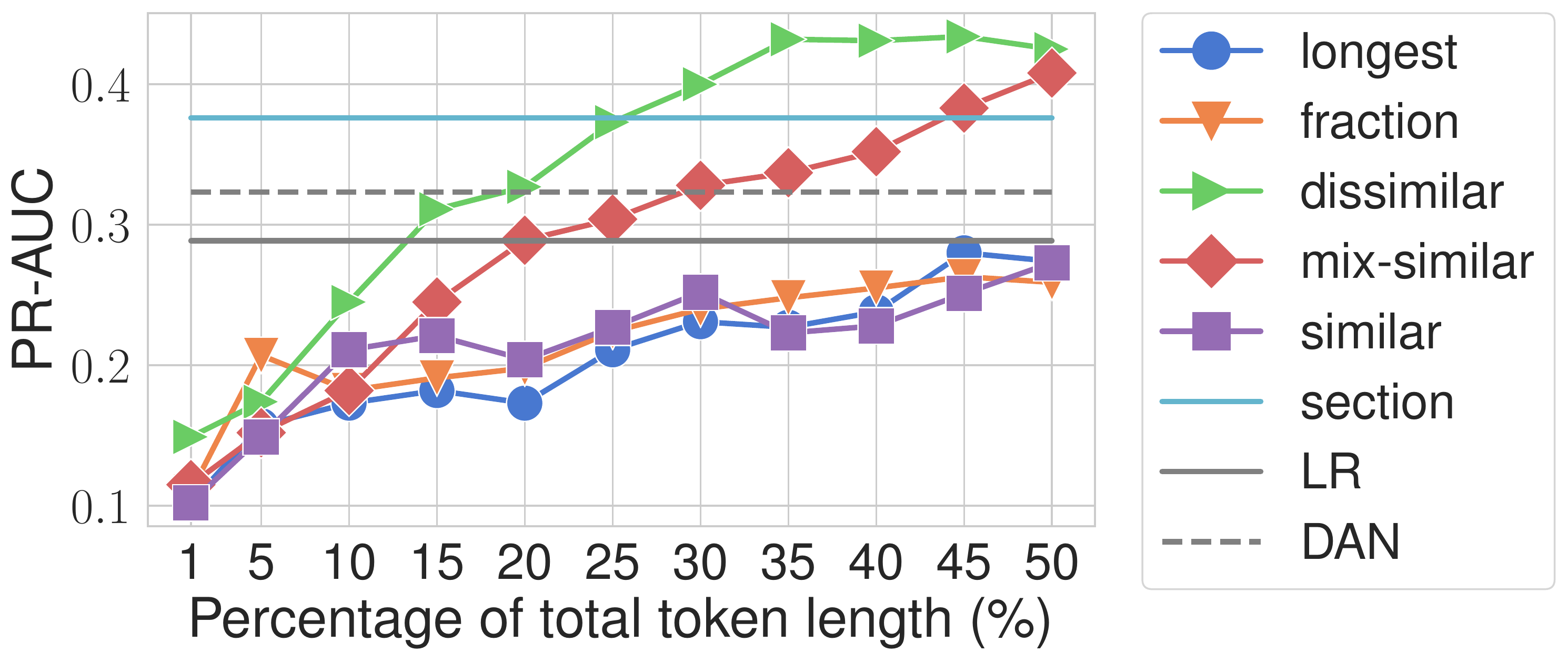}
    \caption{Mortality prediction.}
    \label{fig:heuristics-lr-mortality-main}
  \end{subfigure}
  \caption{Performance of the selected information based on different value functions using the logistic regression (LR) model trained on all notes.
  Despite the distribution shift (selected content is much shorter than the training data, i.e., all notes), the selected information outperforms using all notes with either LR or DAN. 
  }
  \label{fig:heuristics-lr-main}
\end{figure*}
\begin{figure*}
  \centering
  \begin{subfigure}[t]{0.22\textwidth}
    \centering
    \includegraphics[trim=0 0 350 0,clip,width=\textwidth]{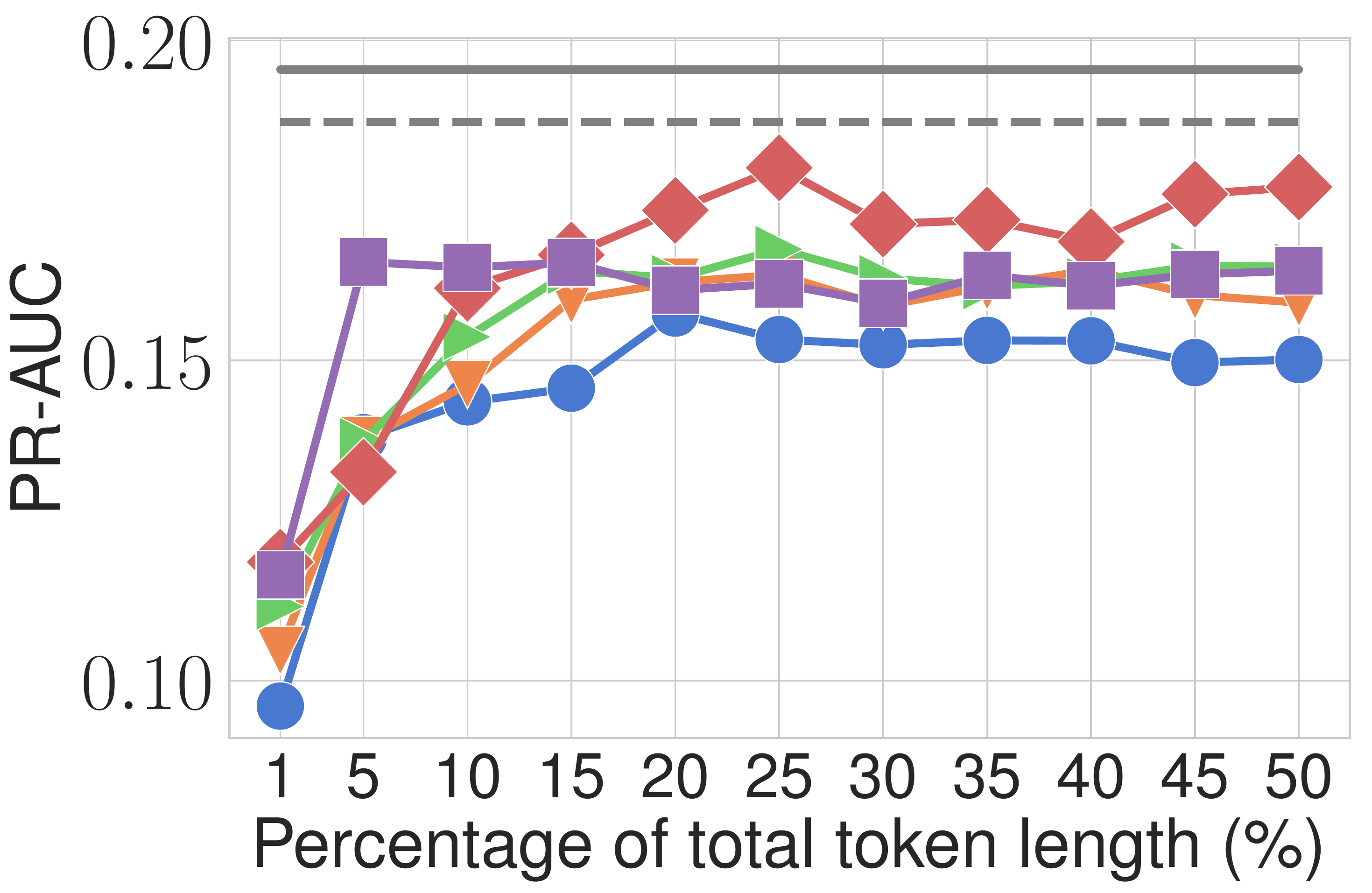}
    \caption{First quartile.}
    \label{fig:q1}
  \end{subfigure}
  \hfill
  \begin{subfigure}[t]{0.22\textwidth}
    \centering
    \includegraphics[trim=0 0 350 0,clip,width=1\textwidth]{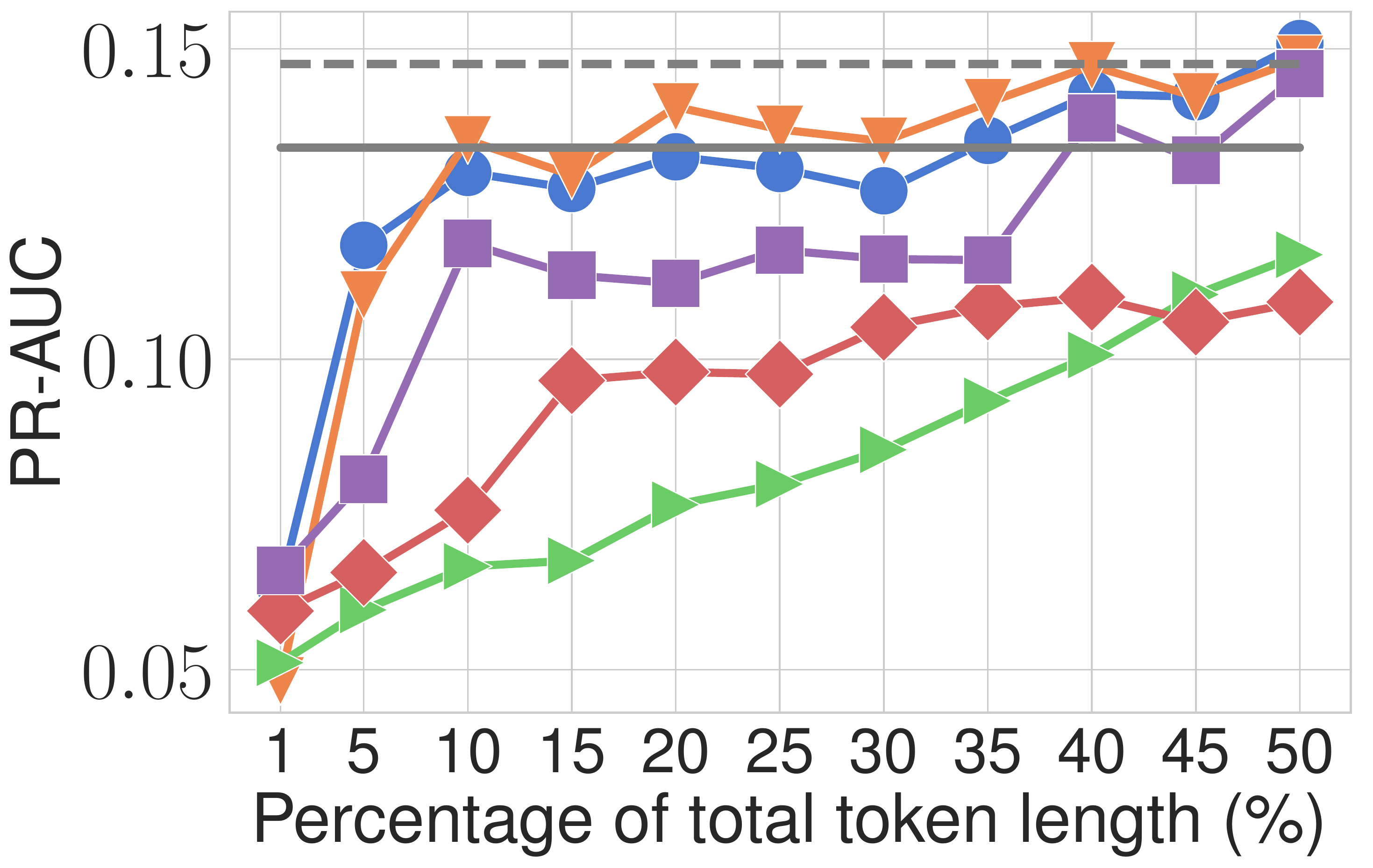}
    \caption{Second quartile.}
    \label{fig:q2}
  \end{subfigure}
  \hfill
  \begin{subfigure}[t]{0.215\textwidth}
    \centering
    \includegraphics[trim=0 0 350 0,clip,width=1\textwidth]{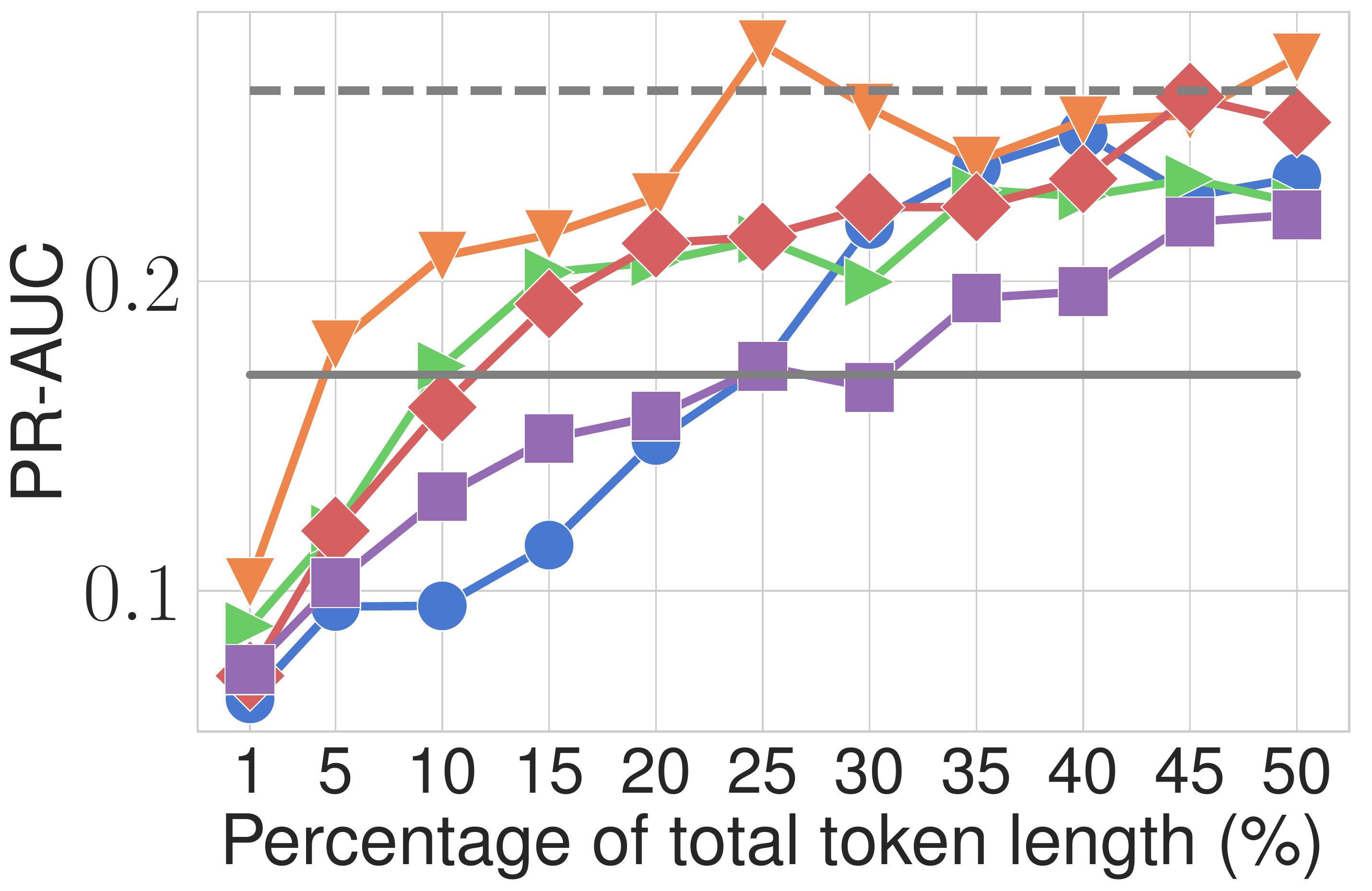}
    \caption{Third quartile.}
    \label{fig:q3}
  \end{subfigure}
  \hfill
  \begin{subfigure}[t]{0.31\textwidth}
    \centering
    \includegraphics[width=1\textwidth]{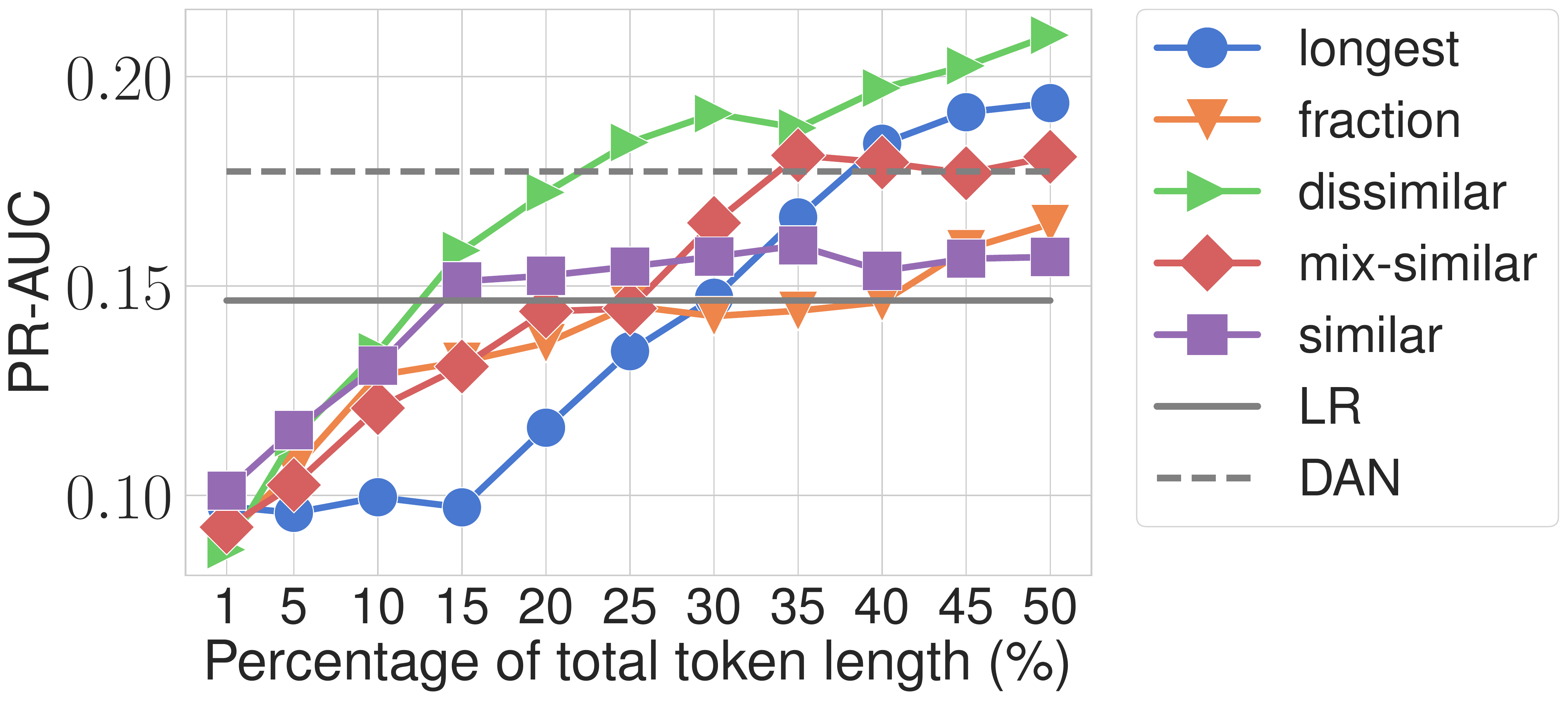}
    \caption{Fourth quartile.}
    \label{fig:q4}
  \end{subfigure}
  \caption{Performance comparison for discharge summaries of different lengths in readmission prediction.
  Selecting valuable information is most useful for the fourth quartile, the longest discharge summaries.
  }
  \label{fig:quartiles}
\end{figure*}

\subsection{Identifying Valuable Chunks of Notes}
\label{sec:heuristics}

Next, we 
zoom into sentences within 
notes to find out which sentences are more valuable, i.e., providing better predictive power using the 
model trained with all notes. 
We choose content from discharge summaries for readmission prediction because they are the most valuable.
For mortality prediction, we select content from the last physician note since they play a similar role as discharge summaries.
To select valuable sentences from $S_i$,
we propose various value functions $V$, and for each $V$, we choose the sentences in $S_i$ that score the highest using $V$ to construct $s_i^V \subset S_i$.
These value functions are our main subject of interest.
We consider the following value functions.
\begin{itemize}[itemsep=0pt, leftmargin=*, topsep=-2pt]
\item \textbf{Longest sentences.}
Intuitively, longer sentences may contain valuable information.
Hence, we use $V_{\operatorname{longest}}(s)=\operatorname{length}(s)$, where $\operatorname{length}$ gives the number of tokens.

\item \textbf{Sentences with highest fractions of medical terms.}
Medical terms are critical for communicating medical information.
We develop a value function based on the fraction of medical terms in a sentence.
Empirically, we observe that fraction alone tends to choose very short sentences, we thus use $V_{\operatorname{frac}}(s) = \frac{\operatorname{medical}(s)}{\operatorname{length}(s)} * \sqrt{\operatorname{length}(s)}$,
where the medical terms come from OpenMedSpel~\cite{OpenMedSpel} and MTH-Med-Spel-Chek~\cite{MTH-Med-Spel-Chek}\footnote{\url{https://github.com/Glutanimate}.}.

\item \textbf{Similarity with previous notes.}
A significant complaint about 
notes is the prevalence of copy-pasting.
We thus develop a value function based on similarity with previous notes.
As discharge summaries are the final 
note within an admission, we compute the max tf-idf similarity of a sentence with all previous notes.\footnote{We also consider average and normalizations to account for the effect of length. We tried to use previous sentences as basic units as well. The results are similar and see the supplementary material for details.}
Specifically, we define 
\mbox{\small $V_{\operatorname{dissimilar}}(s) = -\max_{x_k \in X} \operatorname{cossim}(s, x_k)$}, where $X$ refers to all previous notes:
we find the most similar previous note to the sentence of interest and flip the sign to derive dissimilarity.

Although we hypothesize dissimilar sentences are more valuable due to copy-pasting concerns (i.e., novelty), sentences may also be repeatedly emphasized in notes because they convey critical information. 
We thus also flip $V_{\operatorname{dissimilar}}$ to choose the most similar sentences ($V_{\operatorname{similar}}$) and use $V_{\operatorname{mix}}$ to select half of the most similar and half of the most dissimilar ones.
Similarly, we apply these value functions on the last physician note to select valuable content for mortality prediction. 
\item \textbf{Important Section.}
Finally, physicians do not treat every section in notes equally themselves,
and spend more time on reading the ``Impression and Plan'' section than other sections \citep{brown2014physicians}. 
We use whether a sentence is in this section as our final value function.
This only applies to physician notes.
\end{itemize}

In practice, sentences in medical notes can be very long.
To be fair across different value functions, we truncate the selected sentences to use the same number of tokens with each value function (see the implementation details in the supplementary material).

\begin{table*}[]
  \small
  \centering
  \begin{tabular}{@{}ccl@{}}
  \toprule
  Value Function & Prob. & \multicolumn{1}{c}{Selected Sentences}                                                                        \\ \toprule
  similar        & 0.189       & \begin{tabular}{@{}lcl@{}}Congestive Heart Failure - Systolic and [**Month/Day/Year**] Failure 
    - most recent echo \\ on [**2123-9-3**] with EF 40 \% 
    3. Valvular Disease 
    - Moderate Aortic Stenosis
    - mild\\ - moderate aortic regurgitation 
    - mild - moderate mitral regurgitation 
    4. Coronary Artery Disease\\ - [**2122-11-16**] - s/p BMS to OM2, D1, Left circumflex in [**2122-11-16**] \\for unstable angina and TWI in V2 - 4 - [**2123-5-24**] - NSTEMI s/p cardiac cath\end{tabular}  \\\midrule
  dissimilar     & 0.291        & \begin{tabular}{@{}lcl@{}}No thrills, lifts. BNP elevated though decreased from prior. Please take tiotropium bromide \\ (Spiriva) inhalation twice a day 
  2. Mom[**Name(NI) 6474**] 50 mcg / Actuation Spray ,\\ Non - Aerosol Sig : Two (2) 
  spray Nasal twice a day. Troponins negative x3 sets. PO Q24H\\ (every 24 hours). No S3 or S4. Resp 
  were unlabored, no accessory muscle use. Occupation: \\general surgeon in [**Location (un) 4551**. Abd: Soft, NTND. EtOH: 1 glass of wine or \\
  alcoholic drink /week. + [**4-16**] word dyspnea \end{tabular}    \\\midrule
  mix-similar            & 0.620        & \begin{tabular}{@{}lcl@{}}Congestive Heart Failure - Systolic and [**Month/Day/Year**] Failure 
    - most recent echo \\on [**2123-9-3**] with EF 40\% 
    3. Valvular Disease 
    - Moderate Aortic Stenosis 
    - mild \\- moderate aortic regurgitation 
    - mild - moderate mitral regurgitation 
    4. Coronary Artery Disease \\ \underline{No 
    thrills, lifts. BNP elevated though decreased 
    from prior. Please take tiotropium bromide}\\ \underline{(Spiriva) inhalation twice a 
    day
    2. Mom[**Name (NI) 6474**] 50 mcg / Actuation Spray,}\\ \underline{Non-Aerosol Sig: Two (2) 
    spray Nasal twice a day. Troponins negative x3 sets. PO Q24H}\end{tabular}      \\\bottomrule
  \end{tabular}
  \caption{Example of selected sentences (5\% of tokens) by different value functions from discharge summaries for readmission prediction. 
  This patient was \textbf{readmitted} to the hospital in 30 days after discharge. Underlined sentences in \textit{mix-similar} function come from dissimilar sentences. ``Prob.'' shows the output probability of readmission with the LR model trained on all notes given selected sentences.}
  \label{tab:example}
\end{table*}

\para{Parts of notes can outperform the whole.}
\figref{fig:heuristics-lr-main} shows the test performance of using different value functions to select a fixed percentage of tokens in the discharge summary or the last physician note, compared to using all notes.
The underlying model is the corresponding logistic regression model.
We also show the performance of using all notes with DAN as a benchmark.

Some value functions are able to select valuable information that outperforms using all notes with either logistic regression or DAN.
Interestingly, we find that selected valuable information generally performs better based on the LR model, which seems more robust to distribution shifts than DAN (recall that selected valuable information is much shorter than the expected test set using all notes).

In readmission prediction, medical terms are fairly effective early on, outperforming using all notes with LR, using only 20\% of the discharge summary.
As we include more tokens, %
a mix of similar and dissimilar sentences 
becomes more valuable and  
is eventually comparable with DAN using 45\% of the discharge summary.
Table~\ref{tab:example} presents an example of sentences selected from different value functions in readmission prediction using logistic regression.

In mortality prediction, the advantage of selected valuable information is even more salient.
Consistent with \citet{brown2014physicians}, 
``assessment and plan'' 
is indeed more valuable than the whole note. It alone outperforms both LR and DAN with all notes.
Different from readmission prediction, sentences dissimilar to previous notes are most effective.
The reason might be that 
dissimilar sentences give novel developments in the patient that relate to the impending death.
As structured information dominates 
notes in this task, selected information 
adds little value to structured information
(see 
the supplementary material). 

\para{The effectiveness of value functions varies across lengths.}
To further understand the effectiveness of value functions, we break down \figref{fig:heuristics-lr-readmission-main} based on the length of discharge summaries.
Intuitively, it would be 
harder to select valuable information for short summaries, and \figref{fig:q1} confirms this hypothesis.
In all the other quartiles, a value function is able to select sentences that outperform both LR and DAN using all notes.
The medical terms are most effective in the second and third quartiles.
In the fourth quartile (i.e., the longest discharge summaries), dissimilar content is very helpful,
which likely includes novel perspectives synthesized in discharge summaries. %
These observations resonate with our earlier discussion that dissimilar content contribute novel information.

\subsection{Leveraging Valuable Information}
\label{sec:train_value}

\begin{figure}[t]
\centering
  \includegraphics[width=0.35\textwidth]{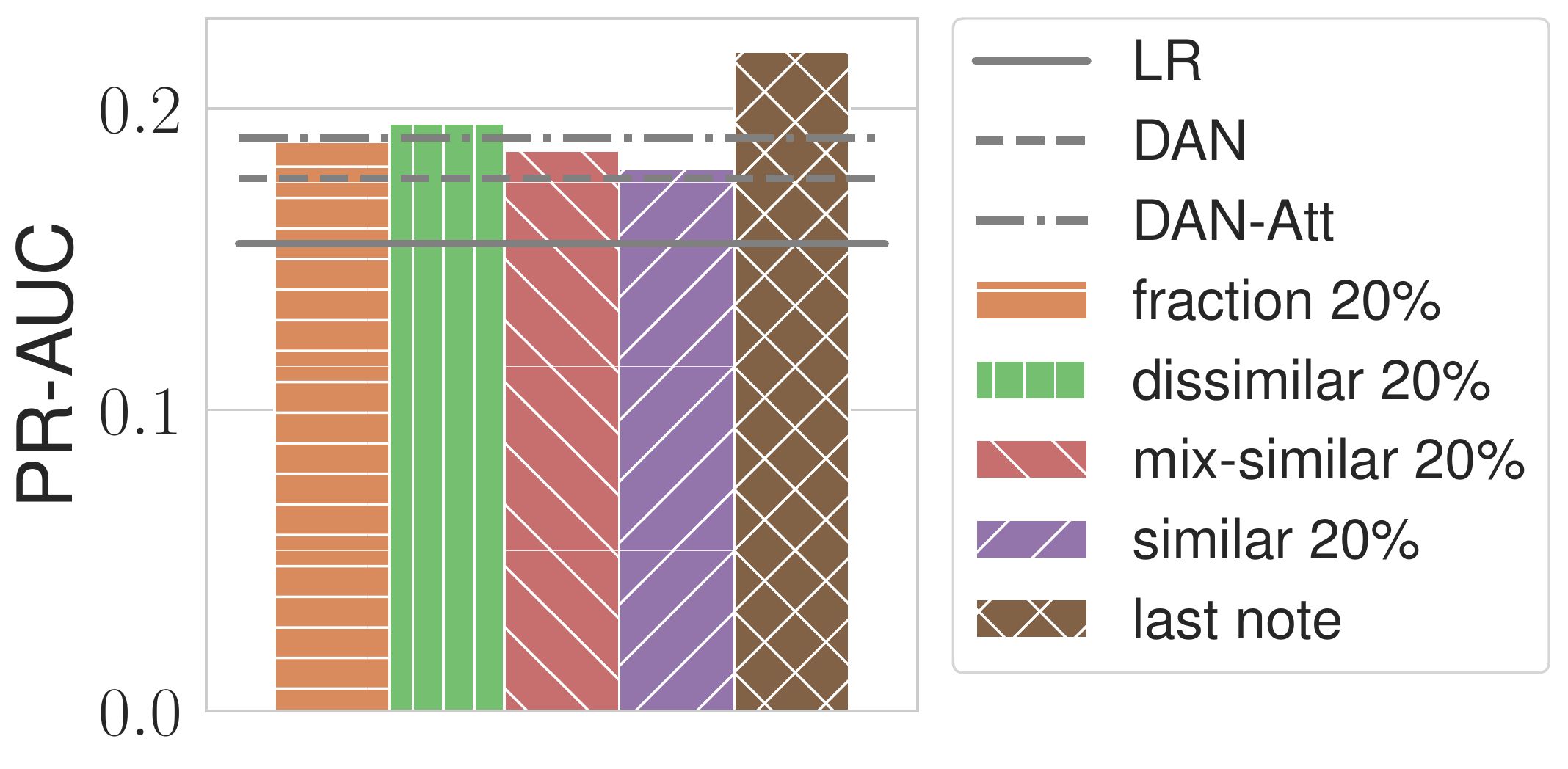}
  \caption{Performance of trained models with selected valuable information (20\% of discharge summaries).
  }
  \vspace{-0.5em}
  \label{fig:retrain}
\end{figure}

Building on the above observations, we leverage the selected valuable information to train models based on only valuable information.
\figref{fig:retrain} shows the performance of these models on readmission prediction.\footnote{This is hard to operationalize for mortality prediction since not all admissions have physician notes.}
Here we include DAN with note-level attention 
(``DAN-Att'') as a model-driven oracle weighted selection approach, although it does not lead to interpretable value functions that can inform caregivers during note-taking. 

First, models trained only using discharge summaries (``last note'') improves the performance over using all notes by 41\% (0.219 vs. 0.155), and outperform DAN and DAN-att as well.
Using medical terms and all types of similarity methods, we can 
outperform using all notes with models only trained on 20\% tokens of discharge summaries, that is, 
6.8\% of all notes. 
Compared to \figref{fig:heuristics-lr-readmission-main}, by focusing exclusively on these selected 20\% of tokens, the model trained with selected dissimilar sentences outperforms logistic regression by 24.3\% (0.194 vs. 0.156) , DAN by 8.2\% (0.194 vs. 0.178), and DAN-Att by 2\% (0.194 vs. 0.190).
We also experiment with ClinicalBERT with a fixed number of tokens (see the supplementary material).
ClinicalBERT provides comparable performance with logistic regression, and %
demonstrates 
similar qualitative trends.

Recall that medical notes dominate structured information for readmission prediction.
It follows that our best performance with 
selected valuable information all outperform the best performance obtained in \secref{sec:overall}.

\section{Related Work}

We summarize additional related work into the following three areas.

\para{Value of medical notes.}
\citet{boag2018s} finds that notes are useful for some classification tasks, e.g., admission type and gender, and reveals that the choice of note representation is crucial for the downstream task.
Prior work also shows that some important phenotypic characteristics can only be inferred from text reports \cite{shivade2014review}. 
For example, \citet{escudie2017novel} observed that 92.5\% of information regarding autoimmune thyroiditis is only presented in text. 
Despite the potential valuable information in medical notes, prior work also points out the redundancy in EHRs.
\citet{cohen2013redundancy} proposed methods to reduce redundant content for the same patient with a summarization-like fingerprinting algorithm, and show improvements in topic modeling. We also discuss the problem of redundancy in notes, but provide a different perspective by probing what type of information is more valuable than others using our 
framework.

\para{NLP for medical notes.} 
The NLP community has worked extensively on medical notes to alleviate information overload, ranging from summarization \citep{mcinerney2020query,liang-etal-2019-novel-system,alsentzer2018extractive} to information extraction \cite{wiegreffe-etal-2019-clinical,zheng2014entity,wang2018clinical}.
For instance, information extraction aims to automatically extract valuable information from existing medical notes.
While our operationalization seems similar, our ultimate goal is to facilitate information solicitation so that medical notes contain more valuable information.
Recently, generating medical notes has attracted substantial interest that might help caregivers record information \citep{liu2018deep,krishna2020generating}, although they do not take into account the value  of information.\\

\para{Predictive tasks with EHRs.}
Readmission prediction and mortality prediction are important tasks that have been examined in a battery of studies \cite{johnson2017reproducibility,ghassemi2014unfolding,purushotham2018benchmarking,rajkomar2018scalable}.
In MIMIC-III, to the best of our knowledge, we have experimented with the most extensive structured variables and as a result, achieved better performance even with simple models.
Other critical tasks include predicting diagnosis codes \citep{ford2016extracting} and length of stay \citep{rajkomar2018scalable}.
We expect information in medical notes to be valued differently in these tasks as well.

\section{Conclusion}

Our results confirm the value of medical notes, especially for readmission prediction.
We further demonstrate that parts can outperform the whole. 
For instance, selected sentences from discharge summaries can better predict future readmission than using all notes and structured variables.
Our work can be viewed as the reverse direction of adversarial NLP \citep{wallace2019universal}:
instead of generating triggers that fool NLP models, we identify valuable information in texts towards enabling humans to generate valuable texts.

Beyond confirming intuitions that ``assessment and plan'' in physician notes is valuable, our work highlights the importance of nursing notes.
Our results also suggest that a possible strategy to improve the value of medical notes is to help caregivers efficiently provide novel content while highlighting important prior information (mixed similarity).
Substantial future work is required to achieve the long-term goal of improving the note-taking process by nudging caregivers towards obtaining and recording valuable information. 

In general, the issue of effective information solicitation has been understudied by the NLP community.
In addition to model advances, we need to develop human-centered approaches to collect data of better quality from people.
As \citet{hartzband2008off} argued, ``as medicine incorporates new technology, its focus should remain on interaction between the sick and healer.''
We hope that our study will encourage studies to understand the interaction process and the note-taking process, beyond understanding the resulting information as a given.
After all, people are at the center of data.

\section{Acknowledgments.} We thank helpful comments from anonymous reviewers.
We thank the MIMIC team for continually providing invaluable datasets for the research community.

\bibliographystyle{acl_natbib}
\bibliography{refs}		%

\appendix

\section{Experiment Setup}

\para{Filtering invalid data.}
We follow the data preprocessing procedure described in 
\citet{Harutyunyan2019}.
Specifically, we first collect 
the event records of a patient by grouping by patient id. Then, we split events of different admissions 
based on admission id.
For all experiments, we eliminate organ donors, i.e., the patients who died already but were readmitted to donate their organs, and the patients who do not have chart event data. Additionally, events missing admission ids are eliminated. 
Note that not all admissions have medical notes. We only select admissions with notes for our experiments.
After filtering out invalid data, we obtain 34,847/37807 patients and 44,055/48,262 admissions for 48 hours/retrospective mortality prediction.

\para{Model Training}
We trained DAN and GRU-D on a single GPU (Nvidia Titan RTX) with PyTorch~\cite{NEURIPS2019_9015}  and we use AdamW~\cite{reddi2019convergence} as an optimizer. Best models are selected based on PR-AUC scores on validation set. 
For model trained with only notes, we set learning rate to 3e-4; for model trained with both types of features or only structured variables, we choose 1e-4 as our learning rate. We train model up to 20 epochs.
We choose the epochs and learning rate that lead to the best performance on the validation set.
\section{Model Details}
\subsection{Deep Averaging Networks (DAN)}
A DAN is similar to a bag-of-word classification model, but the bag-of-words features are replaced by word embeddings. First, we concatenate all notes in an admission as input text and transform input text into a list of token ids $W = \{w_1, w_2, ... , w_P\}$, for which $P$ denotes token length. Then, we obtain word embeddings of each token from a word embedding matrix $M^{V \times D}$ where $D$ is dimension of word embedding. 
After that, we calculate the mean of all word embeddings as the representation $x_{mean}\in R^{D}$ of the input text. Finally, we concatenate $x_{mean}$ and $e_i$ and feed it into final dense layer to obtain the prediction probability output by a softmax function. 

\subsection{DAN with Attention}
Instead of computing averaging word embedding of all notes as a whole, we first generate average word embedding of each note $x_t$ in an admission separately. Then, we compute attention weights and final text representation $x$ as follows:
\begin{gather}
  s_t = Wx_t + b\\
  \alpha_t = \frac{exp(s_t)}{\sum^{T}_{t=1} exp(s_t)} \\
  x = \sum_{t=1}^{T} \alpha_t x_t
\end{gather}
where $T$ denotes number of notes in the admission and $W \in R^{d}$ is a trainable vector.
$x$ is fed into final layer for prediction. 

\subsection{GRU-D}
In contrast to the system described in \citet{che2018recurrent}, we have two types of input features, structured variables and notes, so the input dimension of our model are number of structured variables (767) plus the word embedding dimension (300).
Note that statistical functions of different time windows on structured variables is not applicable because the input of GRU-D should be the content of a single event in the admission.
Also, we do not impute missing value in note representation, because we cannot pre-compute averaged word embeddings across the training set since word embeddings are continually updated during the training process. As shown in Table 2, GRU-D consistently performs worse than logistic regression and DAN.\\
\begin{table*}[]
  \centering
  \begin{tabular}{@{}llcccccc@{}}
    \toprule
                           &     & \multicolumn{2}{c}{LR} & \multicolumn{2}{c}{DAN} & \multicolumn{2}{c}{GRU-D} \\ \midrule
                           &     & PR-AUC    & ROC-AUC    & PR-AUC     & ROC-AUC    & PR-AUC      & ROC-AUC     \\ \midrule
  \multirow{3}{*}{M:24}    & S+N & 0.575     & 0.891      & 0.484      & 0.850      & 0.278       & 0.736       \\
                           & S   & 0.567     & 0.892      & 0.463      & 0.837      & 0.269       & 0.731       \\
                           & N   & 0.288     & 0.754      & 0.323      & 0.767      & 0.15        & 0.585       \\ \midrule
  \multirow{3}{*}{M:48}    & S+N & 0.558     & 0.903      & 0.520      & 0.888      & 0.254       & 0.764       \\
                           & S   & 0.547     & 0.902      & 0.519      & 0.890      & 0.268       & 0.774       \\
                           & N   & 0.292     & 0.794      & 0.341      & 0.810      & 0.105       & 0.536       \\ \midrule
  \multirow{3}{*}{M:retro} & S+N & 0.927     & 0.983      & 0.902      & 0.978      & 0.684       & 0.912       \\
                           & S   & 0.921     & 0.982      & 0.892      & 0.976      & 0.714       & 0.923       \\
                           & N   & 0.745     & 0.935      & 0.816      & 0.953      & 0.319       & 0.752       \\ \midrule
  \multirow{3}{*}{Readmission} & S+N & 0.156     & 0.714      & 0.189      & 0.741      & 0.132       & 0.619       \\
                           & S   & 0.150     & 0.699      & 0.144      & 0.682      & 0.112       & 0.606       \\
                           & N   & 0.155     & 0.730      & 0.176      & 0.744      & 0.085       & 0.552       \\ \bottomrule
  \end{tabular}
  \caption{Results of PR-AUC/ROC-AUC scores on LR/DAN/GRU-D models in readmission prediction  and mortality prediction (``M'')  tasks. S denotes structured variables and N is notes.}
  \label{tab:my-table}
  \end{table*}

\section{Pairwise Comparison with DAN}
\begin{figure*}[!t]
  \centering
  \begin{subfigure}[t]{0.35\textwidth}
    \centering
    \includegraphics[width=\textwidth]{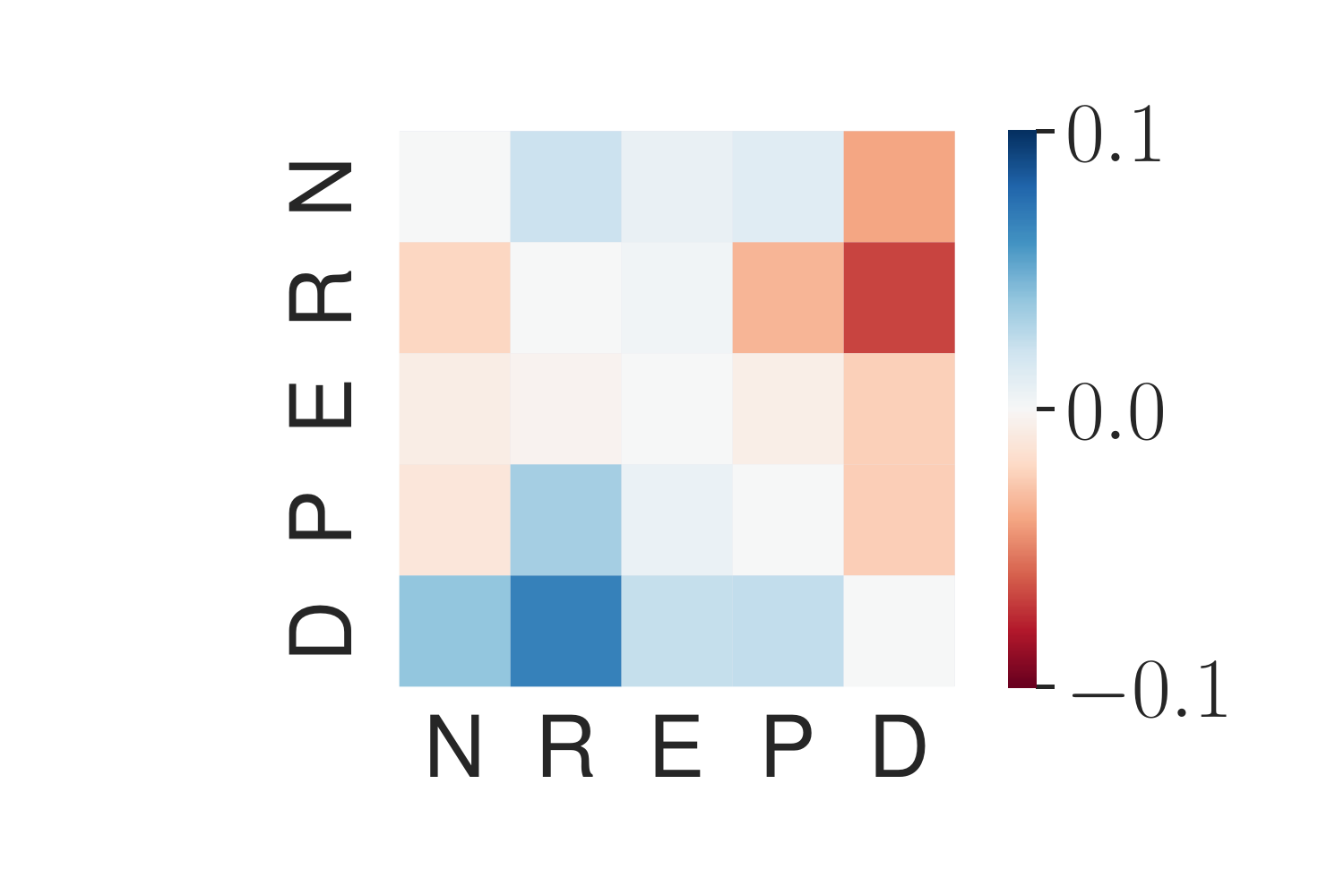} 
    \caption{Readmission prediction}
    \label{fig:pairwise-readmission}
  \end{subfigure}
  \hfill
  \begin{subfigure}[t]{0.35\textwidth}
    \centering
    \includegraphics[width=\textwidth]{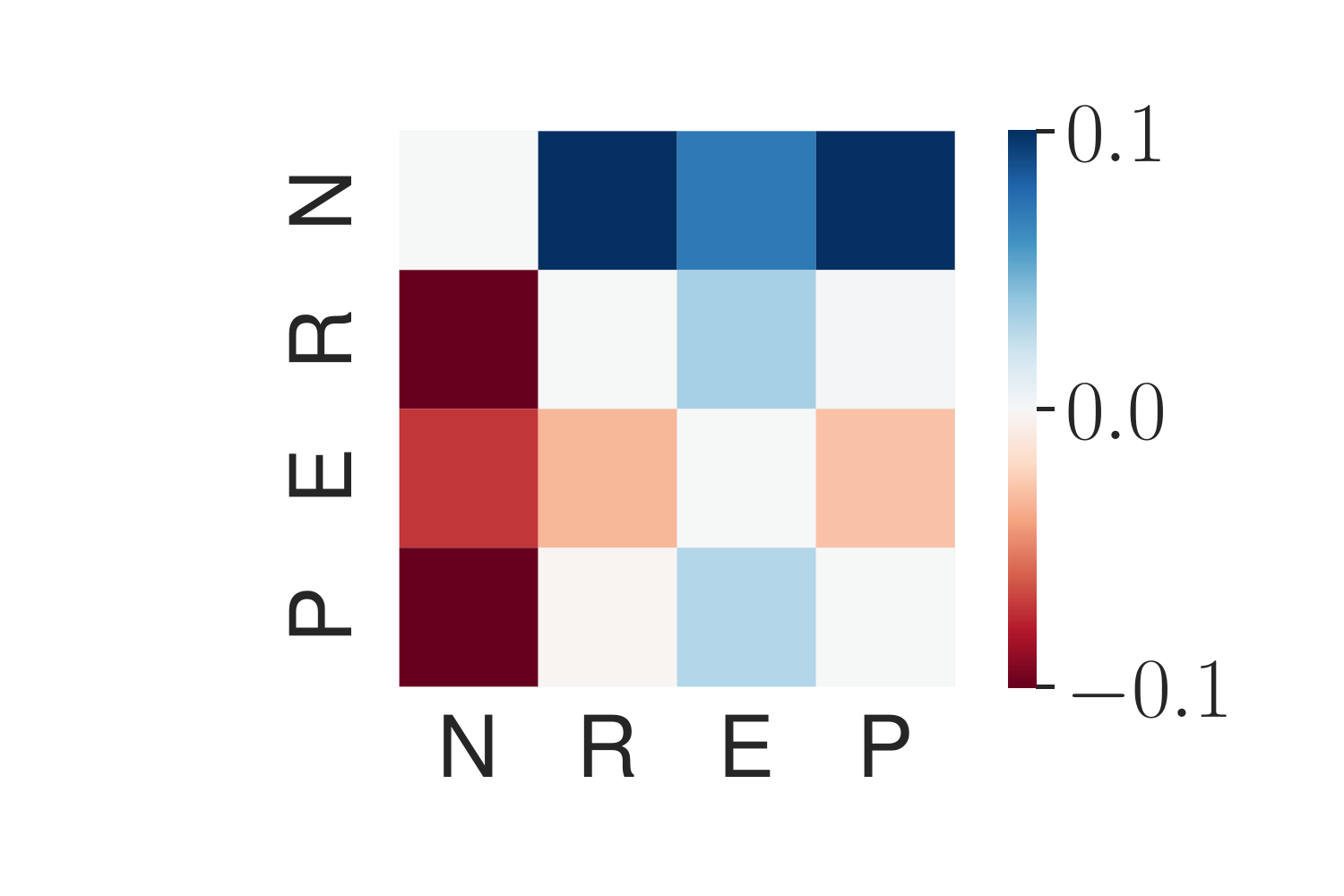}
    \caption{Mortality prediction (24 hrs)}
    \label{fig:pairwise-mortality}
  \end{subfigure}
  \hfill
  \begin{subtable}{0.27\textwidth}
  \vspace{-1in}
  \begin{tabular}{l}
  \toprule
  P: Physician notes \\
  N: Nursing notes \\
  D: Discharge summary\\
  R: Radiology reports \\
  E: ECG reports \\
  \bottomrule
  \end{tabular}
  \end{subtable}
  \caption{Pairwise comparisons between different types of notes on DAN (each grid shows $\operatorname{PR-AUC}(f_{\text{all}}(s_{t_{\operatorname{row}}}), y) - \operatorname{PR-AUC}(f_{\text{all}}(s_{t_{\operatorname{column}}}), y))$.
  To account for the differences in length, we subsample two types of notes under comparison to be the same length and report the average values of 10 samples.
  Discharge summaries dominate all other types of notes in readmission prediction, while nursing notes are most useful for mortality prediction.
  }
  \label{fig:pairwise-main-dan}
  \end{figure*}

Results of pairwise comparison with DAN are similar with results with logistic regression (LR). \figref{fig:pairwise-main-dan} shows that discharge summaries dominate other types of notes for readmission prediction. In mortality prediction, nursing notes are the most useful notes.

\section{Similarity Computation}
We have tried to compute the tf-idf similarity of a sentence in the last note with all previous sentences instead of notes, and results are similar.

\begin{figure*}[!t]
  \centering
  \begin{subfigure}[t]{0.33\textwidth}
    \centering
    \includegraphics[width=1\textwidth]{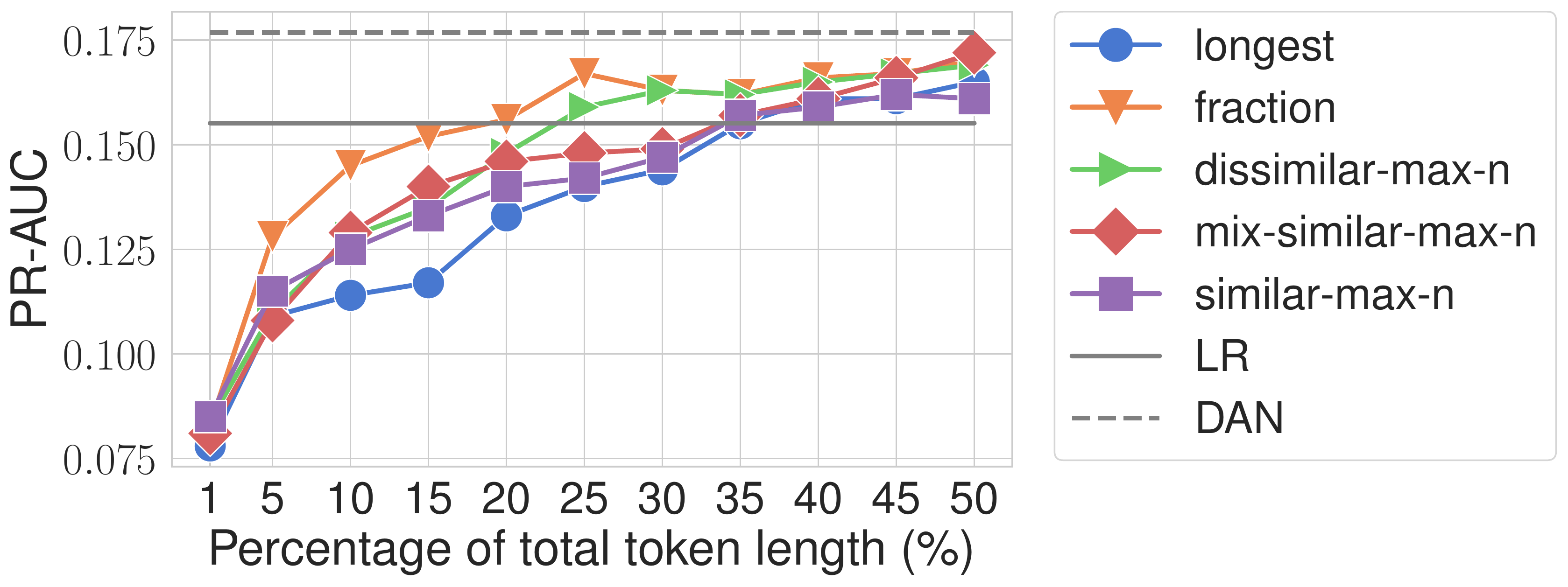}
    \caption{Max norm.}
    \label{fig:heuristics-dan}
  \end{subfigure}
  \begin{subfigure}[t]{0.325\textwidth}
    \centering
    \includegraphics[width=1\textwidth]{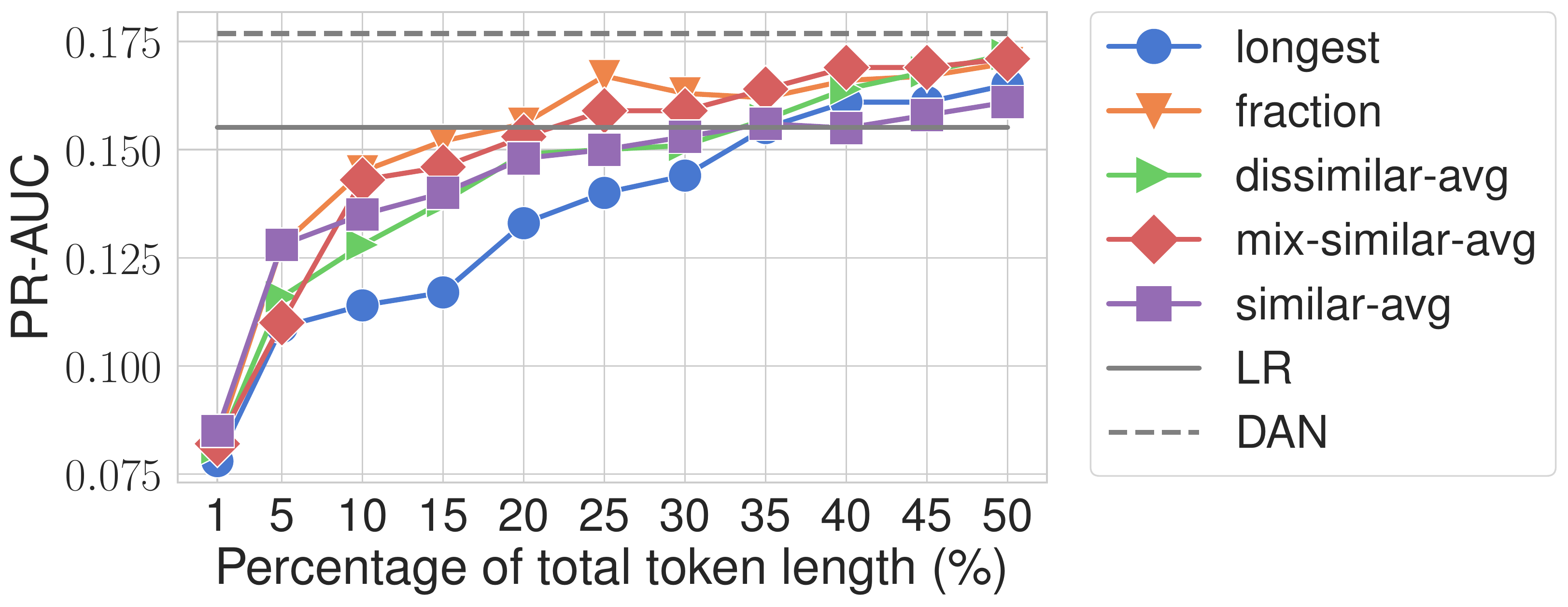}
    \caption{AVG.}
    \label{fig:heuristics-lr}
  \end{subfigure}
    \begin{subfigure}[t]{0.33\textwidth}
    \centering
    \includegraphics[width=1\textwidth]{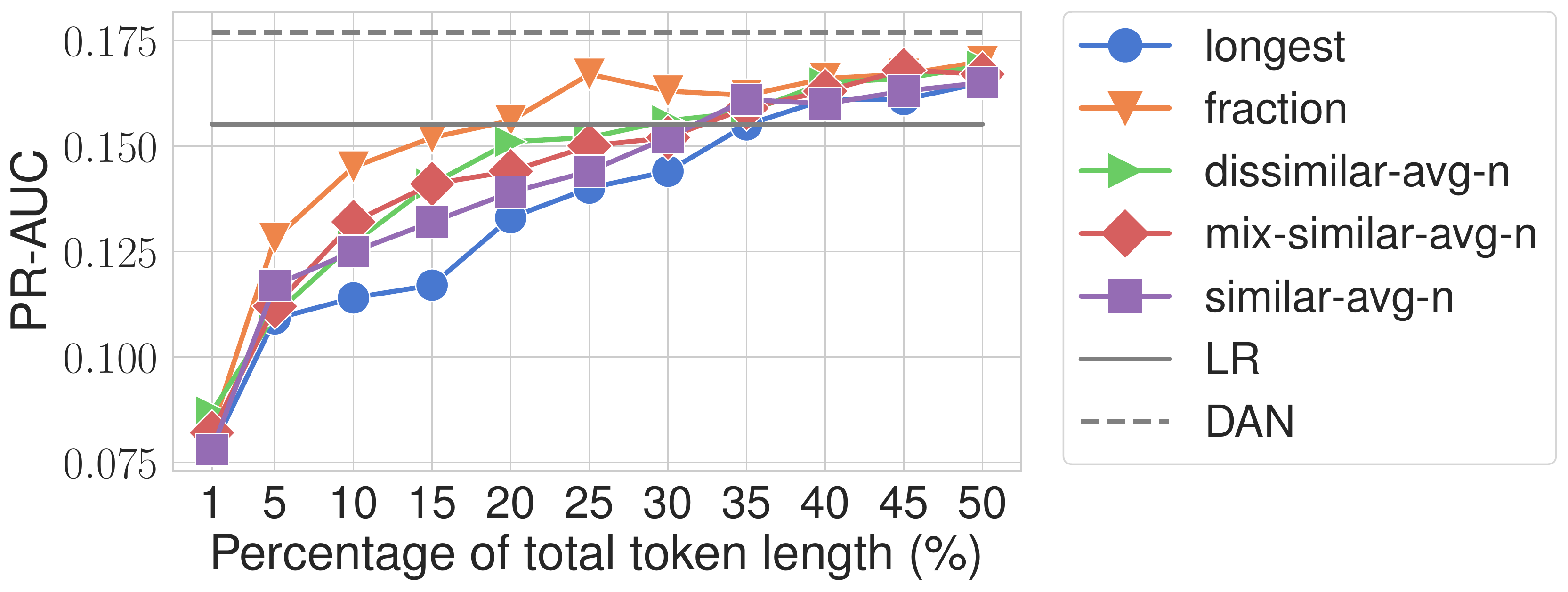}
    \caption{AVG norm.}
    \label{fig:heuristics-lr}
  \end{subfigure}

  \caption{PRAUC scores of other similarity value functions for readmission prediction on LR with different percentages of tokens.}
  \label{fig:similarity-LR-readmission}
\end{figure*}

\begin{figure*}[!t]
  \centering
  \begin{subfigure}[t]{0.33\textwidth}
    \centering
    \includegraphics[width=1\textwidth]{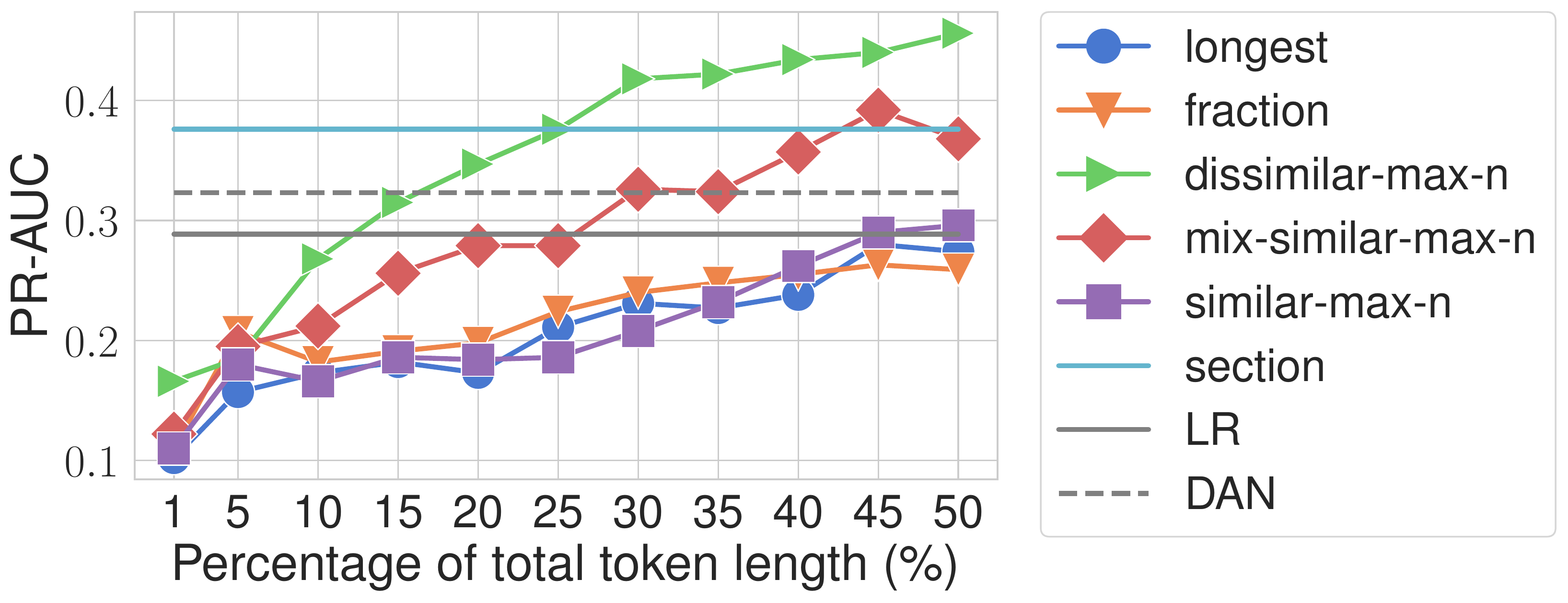}
    \caption{Max norm.}
    \label{fig:heuristics-dan}
  \end{subfigure}
  \hfill
  \begin{subfigure}[t]{0.32\textwidth}
    \centering
    \includegraphics[width=1\textwidth]{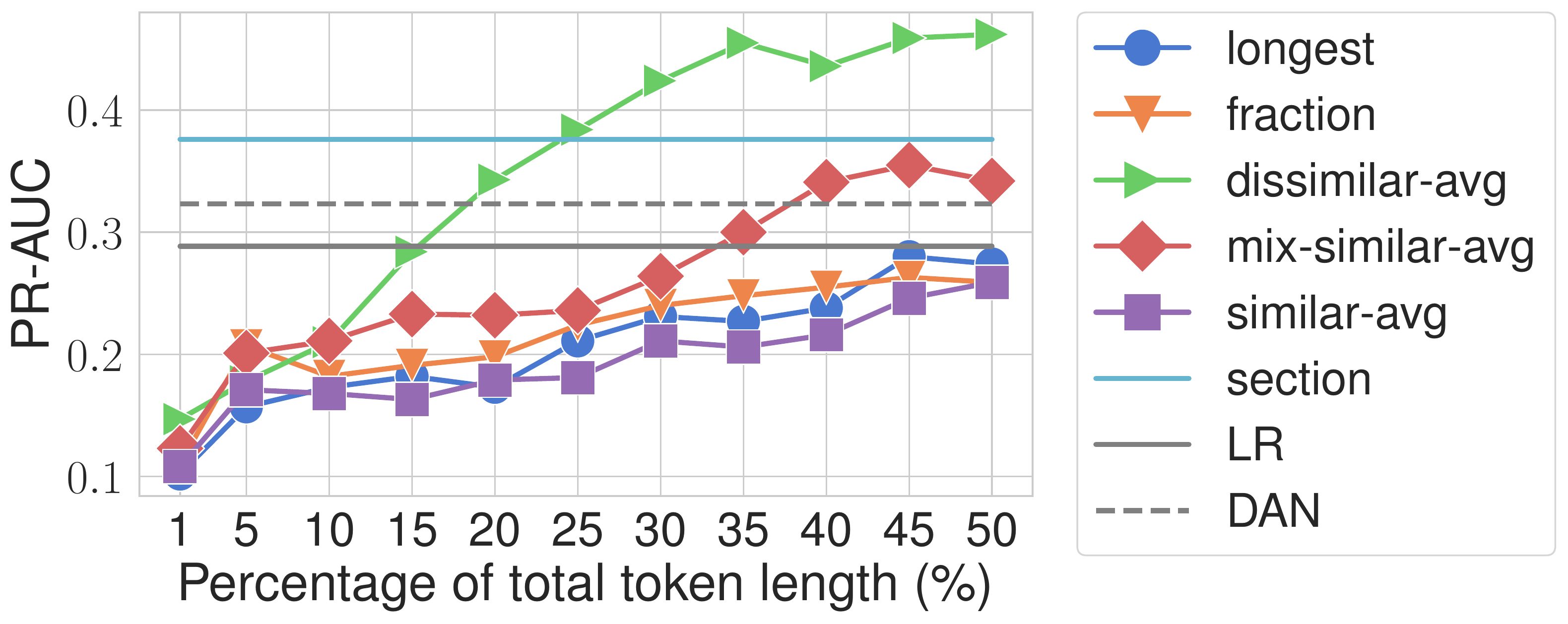}
    \caption{AVG.}
    \label{fig:heuristics-lr}
  \end{subfigure}
  \hfill
  \begin{subfigure}[t]{0.33\textwidth}
    \centering
    \includegraphics[width=1\textwidth]{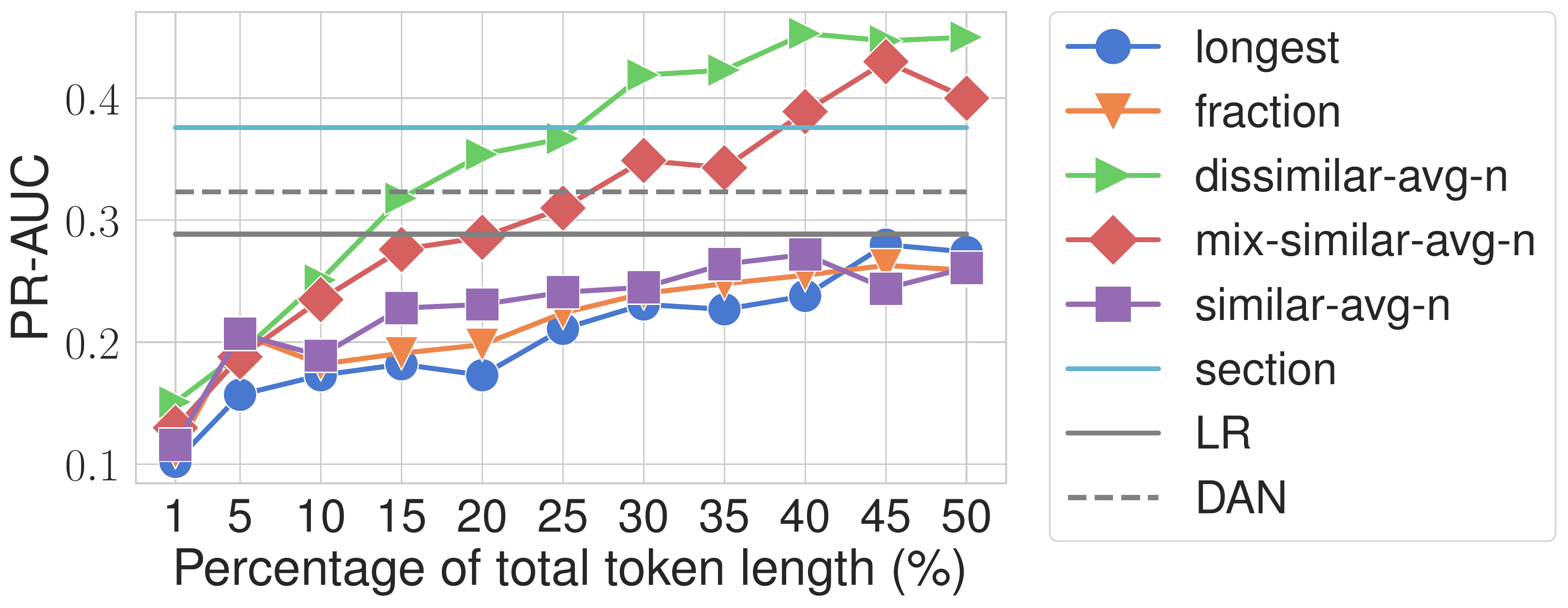}
    \caption{AVG norm.}
    \label{fig:heuristics-lr}
  \end{subfigure}
  \caption{PRAUC scores of other similarity value functions for mortality prediction with 24 hours period on LR with different percentages of tokens.}
  \label{fig:similarity-all_but_discharge-mortality}
\end{figure*}

Besides max similarity value function, we also conduct experiments on average similarity with or without length normalization.
\begin{flalign*}
  V_{sim\_max}(s) &= \max_{x_k\in X}\: cossim(s, x_k)\\
  V_{sim\_avg}(s) &= \small{\frac{1}{k-1}\sum_{k=1}^{K-1}} cossim(s, x_k)\\
  V_{sim\_max\_n}(s) &=\\
   \max_{x_k\in X}\;& cossim(s, x_k)* \sqrt{\operatorname{length}(s)}\\
  V_{sim\_avg\_n}(s) &= \\
  \small{\frac{1}{k-1}\sum_{k=1}^{K-1}}\; &cossim(s, x_k)* \sqrt{\operatorname{length}(s)}
\end{flalign*}
where $K$ denotes number of notes in the admission.
The value functions for dissimilar sentences are as follows:
\begin{flalign*}
  V_{dis\_max}(z) &= -V_{sim\_max}(z)\\
  V_{dis\_avg}(z) &= -V_{sim\_avg}(z)\\
  V_{dis\_max\_n}(z) &= -V_{sim\_max\_n}(z)\\
  V_{dis\_avg\_n}(z) &= -V_{sim\_avg\_n}(z)
\end{flalign*}
While compute similarity, we force selected sentences to have at least five tokens to prevent super short sentences. Results in \figref{fig:similarity-LR-readmission} and \figref{fig:similarity-all_but_discharge-mortality} show no significant difference between max and other similarity methods.\\
\para{Truncate at a given percentage.} To make a fair comparison, we first calculate the total number of tokens from selected sentences which are sorted by scores from a value function. Then, we compute number of tokens given a percentage as $n = n_{total}*\operatorname{percentage}$ and we keep pushing sentences to the output list with descending order of scores until the number of tokens in output list exceed $n$. Last, we truncate exceeding tokens in last selected sentence to obtain final output sentences.

\begin{figure}
  \centering

    \includegraphics[width=1\linewidth]{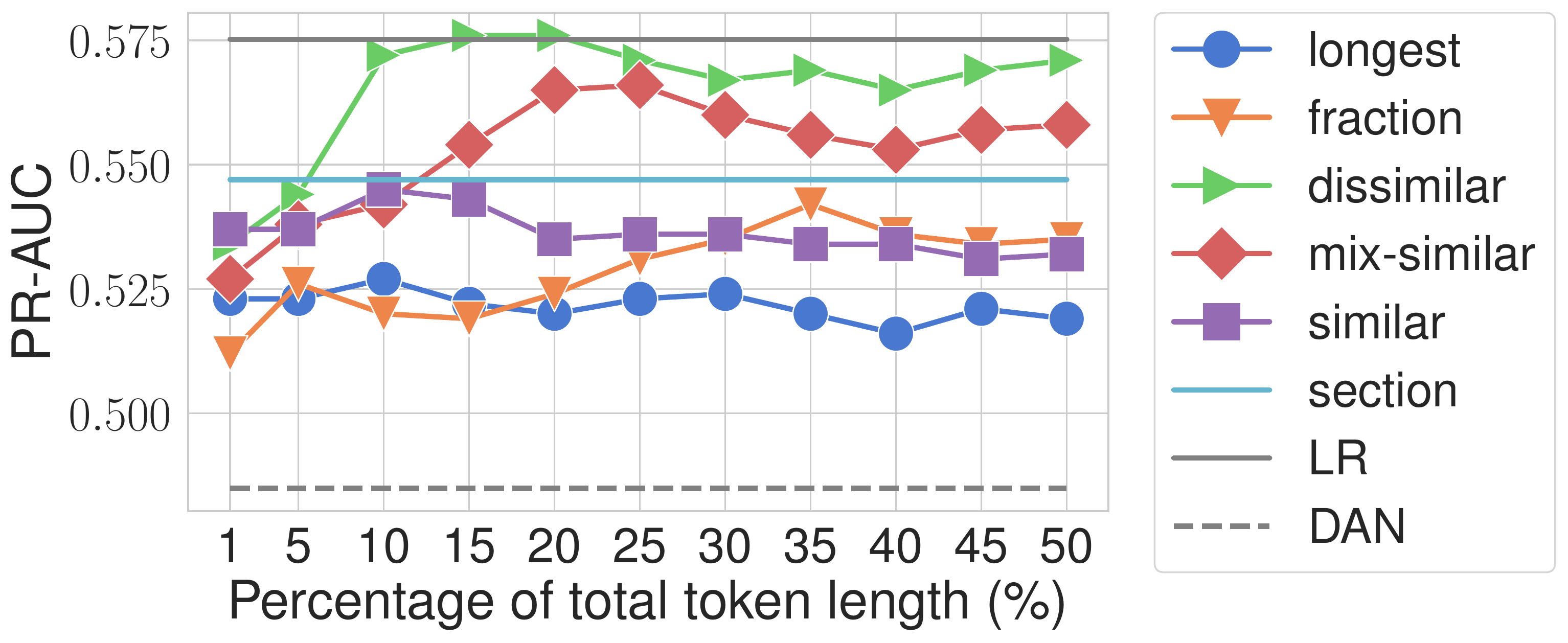}

  \caption{ PRAUC scores of value functions for mortality prediction (24 hours) on LR with different \textbf{percentages of token length} using both structured information and notes. Since structured variables alone already dominate in this tasks, notes do not add additional predictive values to the results.
  }
  \label{fig:structured-dominate}
\end{figure}

\para{Structured information dominates mortality prediction task.} 
As shown in \figref{fig:structured-dominate}, logistic regression already performs well with structured variables alone in mortality prediction. 
Adding part of notes does not add much predictive value over using all notes. 

\begin{figure*}[!h]
  \centering
  \begin{subfigure}[t]{0.45\textwidth}
    \centering
    \includegraphics[trim=0 0 0 0,clip,width=.9\textwidth]{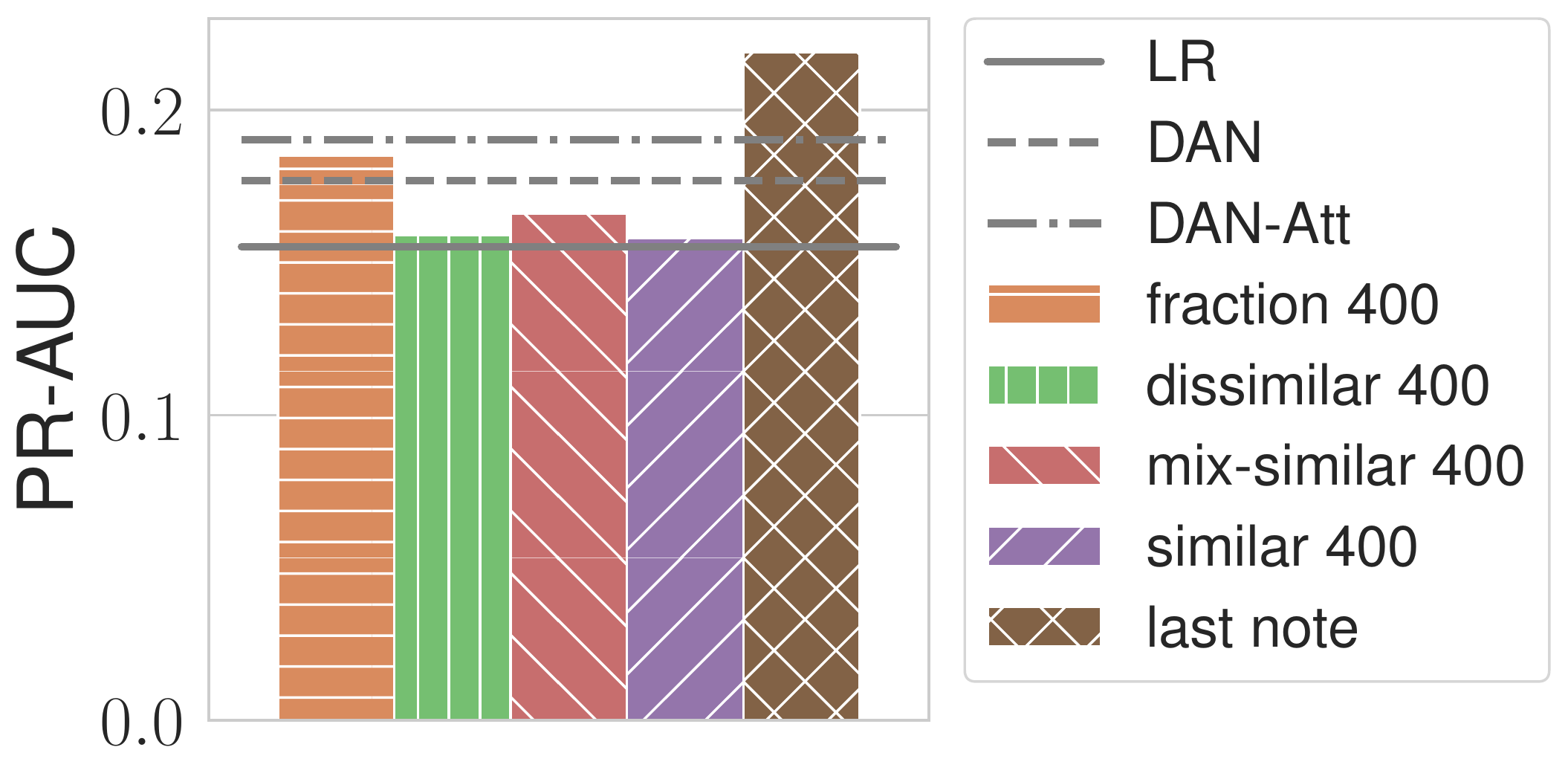}
    \caption{Logistic regression.}
  \end{subfigure}
  \hfill
  \begin{subfigure}[t]{0.45\textwidth}
    \centering
    \includegraphics[trim=0 0 0 0,clip,width=.9\textwidth]{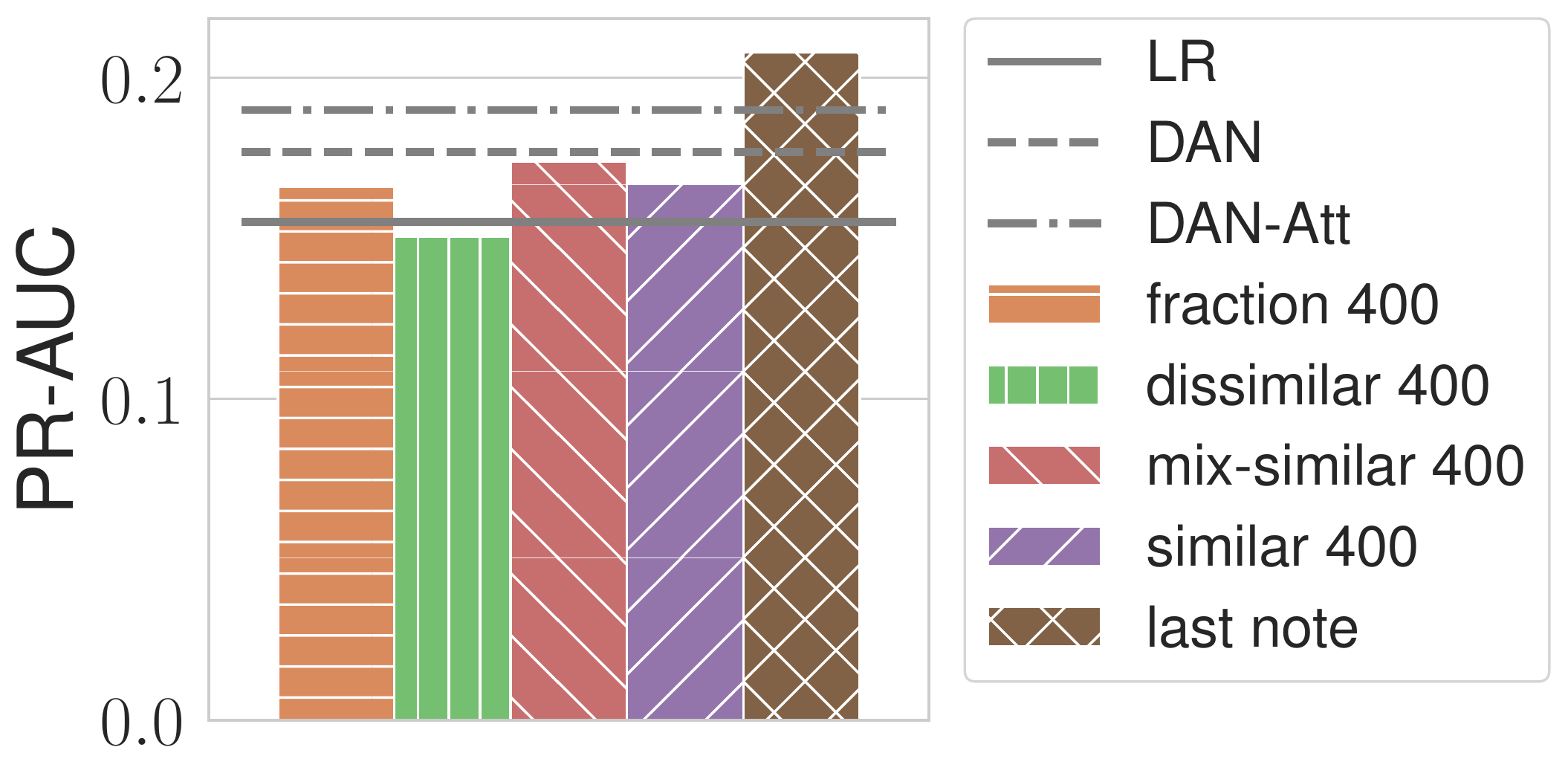}
    \caption{ClinicalBERT}
  \end{subfigure}
    \caption{Performance of trained models with selected valuable information (400 tokens) on logistic regression and ClinicalBert.
    ClinicalBERT does not necessarily provide better performance than logistic regression.
    The mix-similar method is the best among three similarity methods. It is different from re-training models based on \textbf{percentage} of tokens where dissimilar sentences provide highest predictive values.
    }
    \label{fig:bert-retrain}
  \end{figure*}
  \section{Leveraging Valuable Information : Clinical-BERT}
  Since ClinicalBERT has input length limitation of 512 sub-word tokens, we fine-tune ClinicalBert with sentences selected by proposed value functions where sentences are truncated at 400 tokens.
  Note that truncation based on number of tokens instead of percentage of tokens will drastically reduce long records to too little information, and vice versa. 
  As shown in \figref{fig:bert-retrain}, fine-tuned ClinicalBert based on selected sentences has similar performance with re-trained logistic regression model. 
\begin{figure*}[!t]
  \centering

  \large{In-hospital Mortality Prediction (48 hours)}\\
  \begin{subfigure}[t]{0.24\textwidth}
      \centering
      \includegraphics[trim=0 0 1020 0,clip,width=0.95\textwidth]{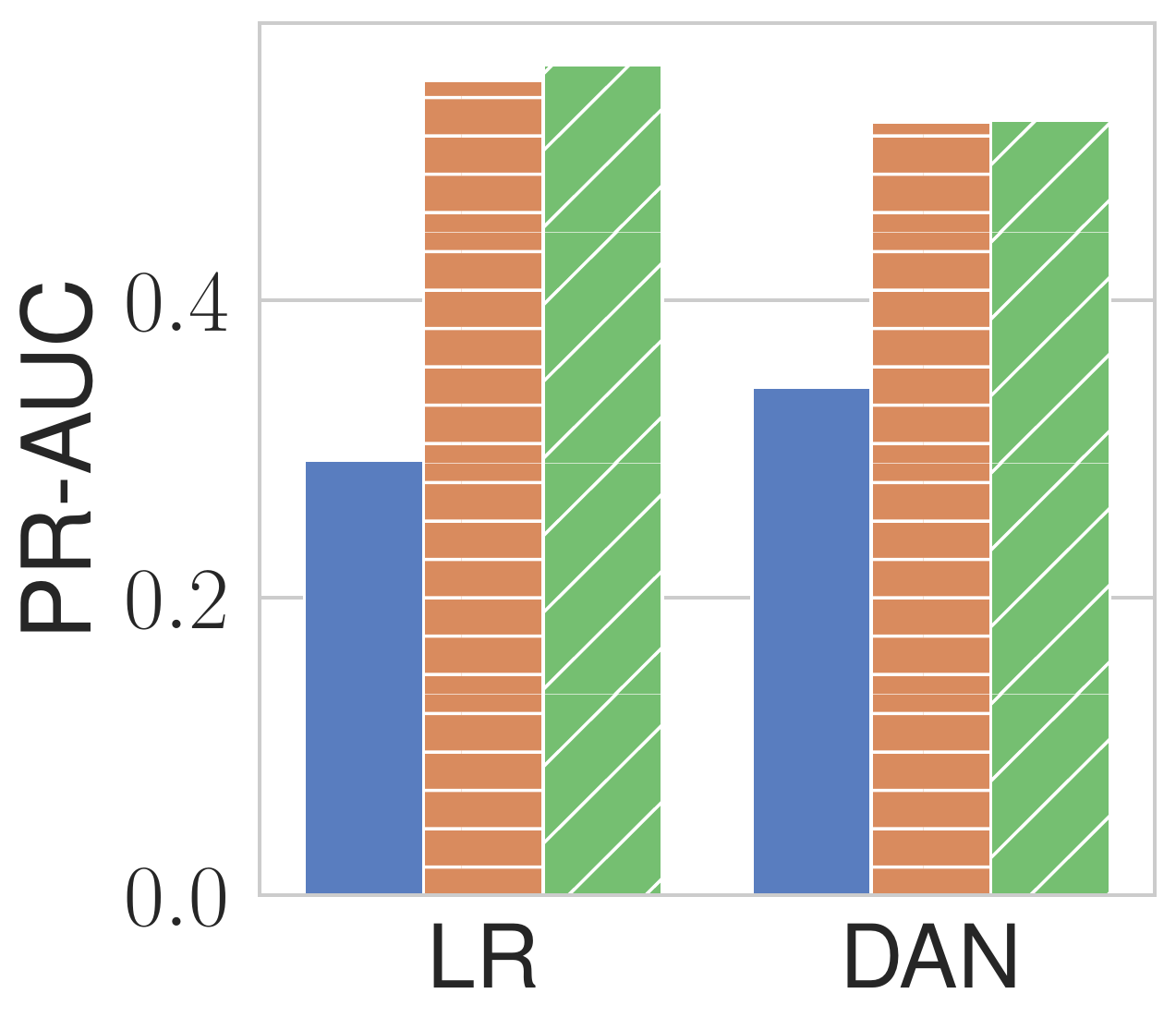}
      \caption{PR-AUC.}
      \label{fig:prauc-mortality}
  \end{subfigure}
  \begin{subfigure}[t]{0.24\textwidth}
      \centering
      \includegraphics[trim=0 0 1020 0,clip,width=0.99\textwidth]{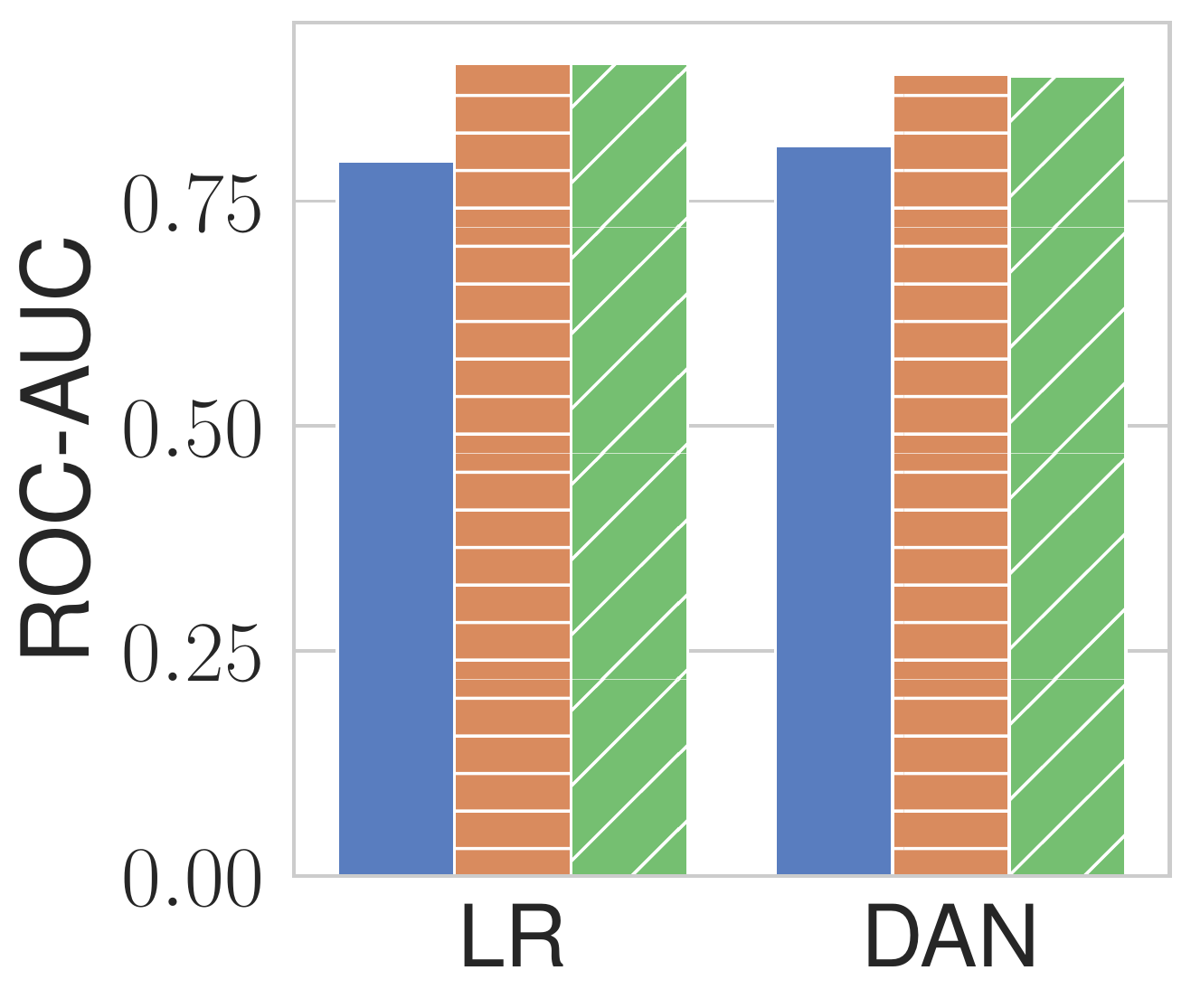}
      \caption{ROC-AUC.}
      \label{fig:rocauc-mortality}
  \end{subfigure}
  \begin{subfigure}[t]{0.24\textwidth}
      \centering
      \includegraphics[trim=0 0 1020 0,clip,width=0.99\textwidth]{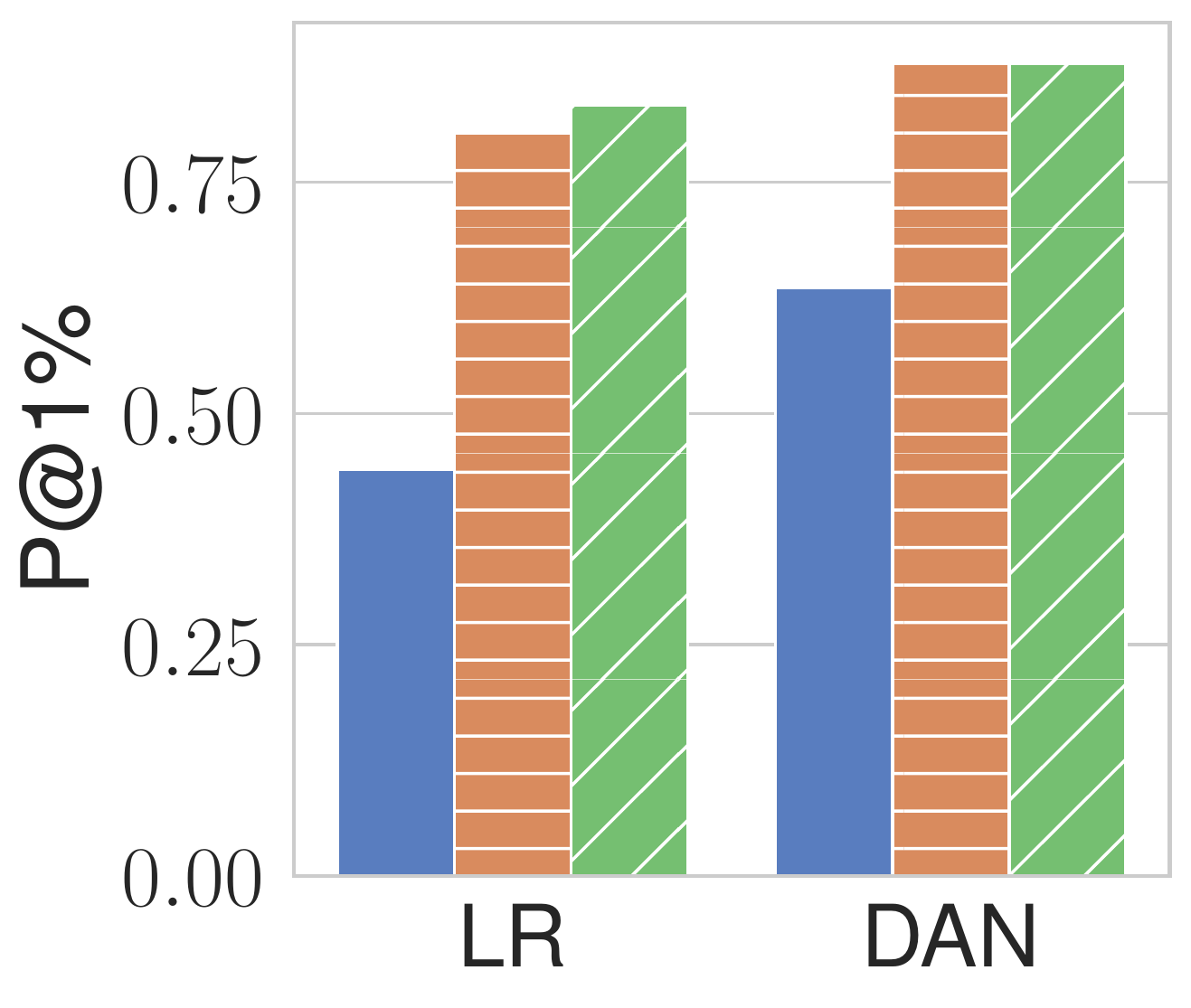}
      \caption{Precision at 1\%.}
      \label{fig:p1-mortality}
  \end{subfigure}
  \begin{subfigure}[t]{0.24\textwidth}
      \centering
      \includegraphics[trim=0 0 1020 0,clip,width=0.95\textwidth]{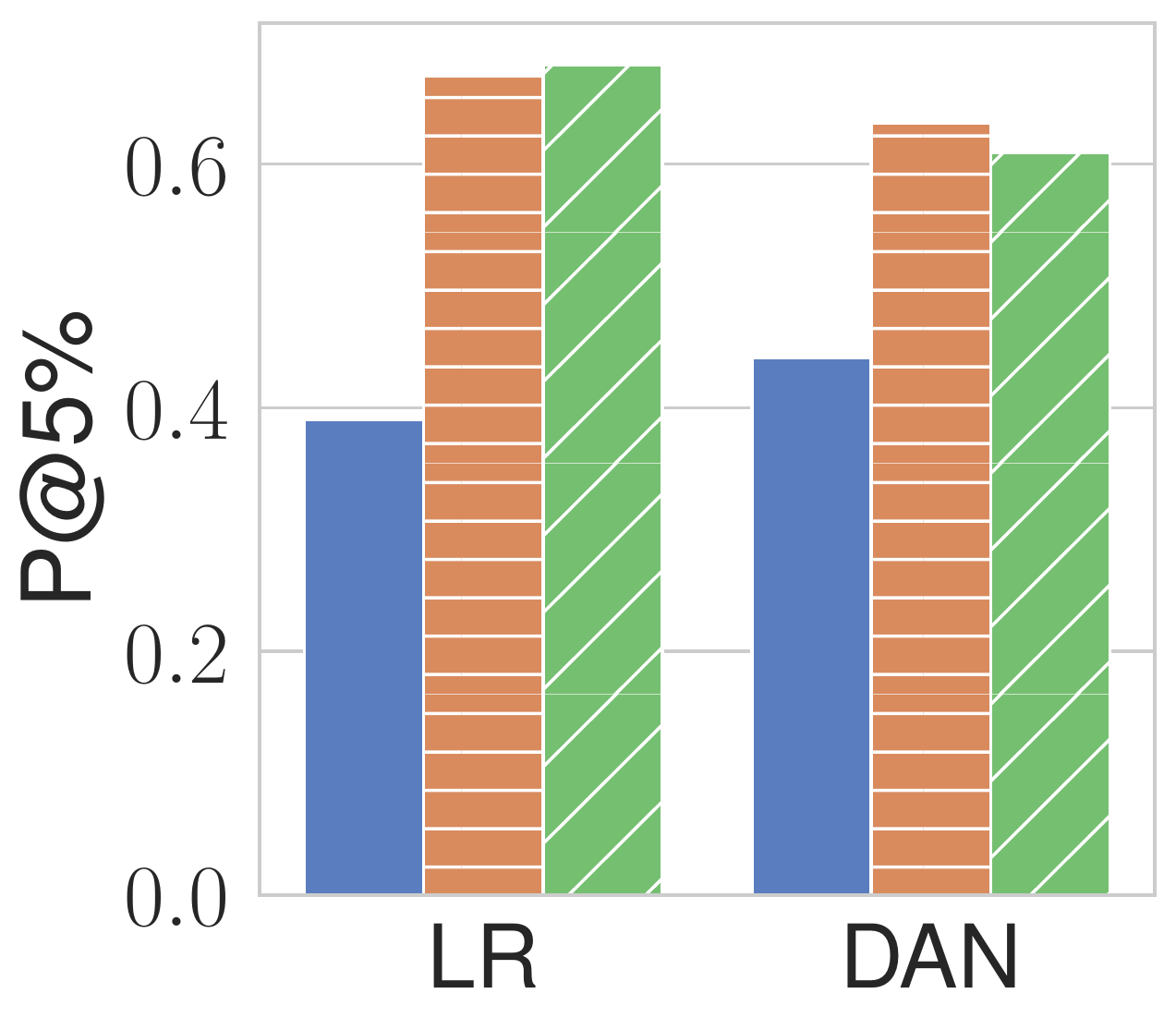}
      \caption{Precision at 5\%.}
      \label{fig:p5-mortality}
  \end{subfigure}
  \large{In-hospital Mortality Prediction (retrospective)}\\
  \begin{subfigure}[t]{0.24\textwidth}
    \centering
    \includegraphics[trim=0 0 1020 0,clip,width=0.95\textwidth]{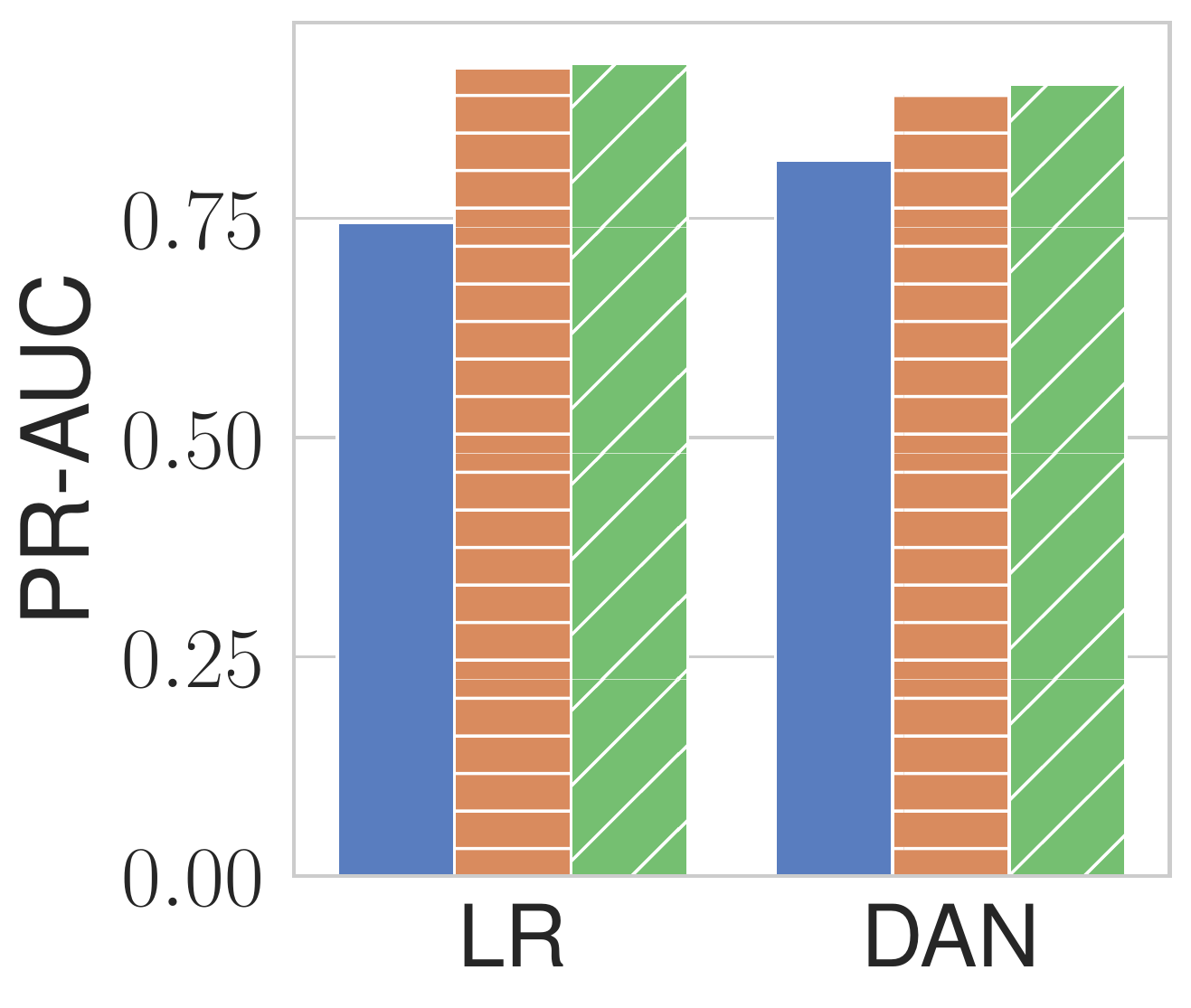}
    \caption{PR-AUC.}
    \label{fig:prauc-mortality}
\end{subfigure}
\begin{subfigure}[t]{0.24\textwidth}
    \centering
    \includegraphics[trim=0 0 1020 0,clip,width=0.99\textwidth]{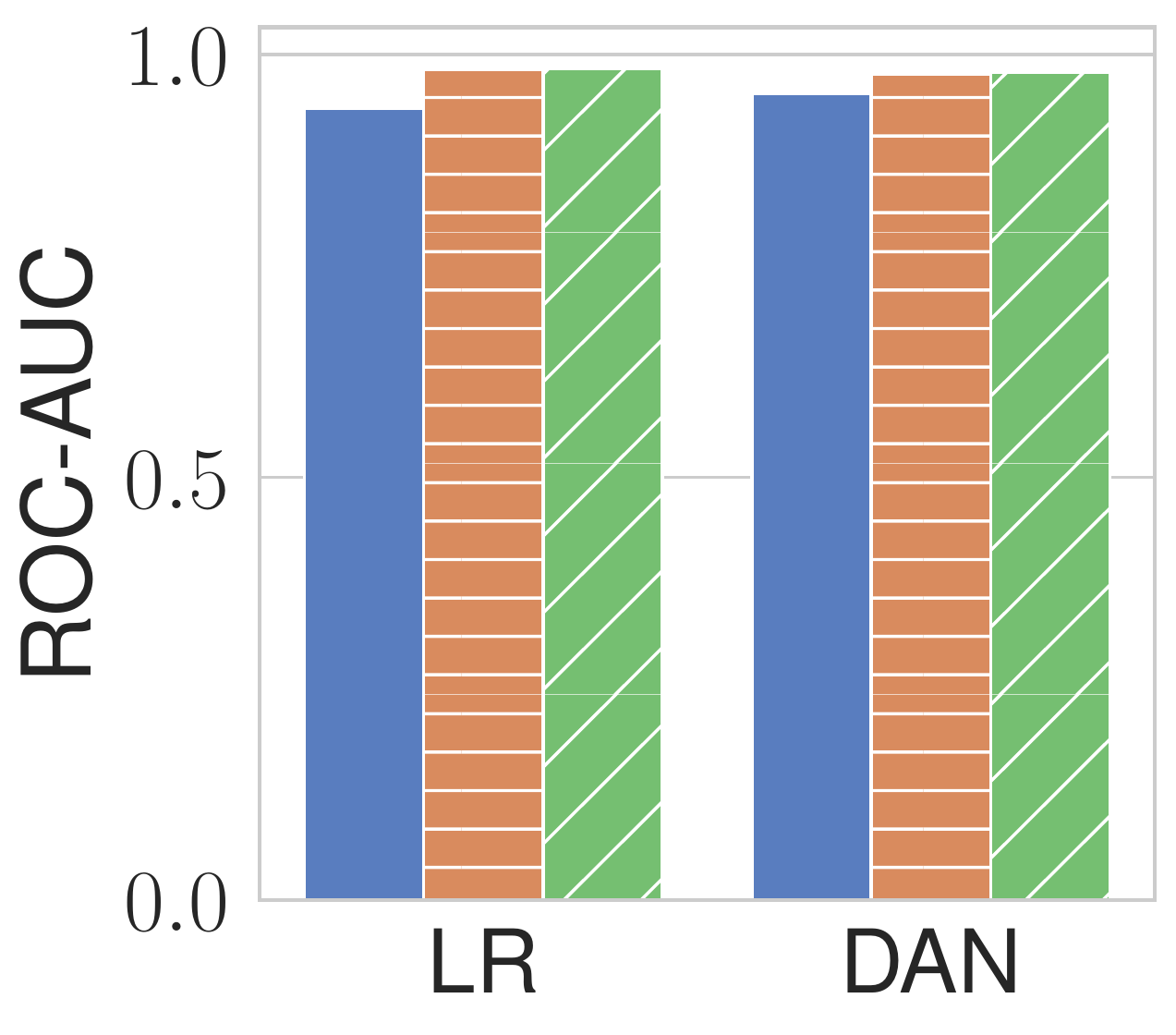}
    \caption{ROC-AUC.}
    \label{fig:rocauc-mortality}
\end{subfigure}
\begin{subfigure}[t]{0.24\textwidth}
    \centering
    \includegraphics[trim=0 0 1020 0,clip,width=0.99\textwidth]{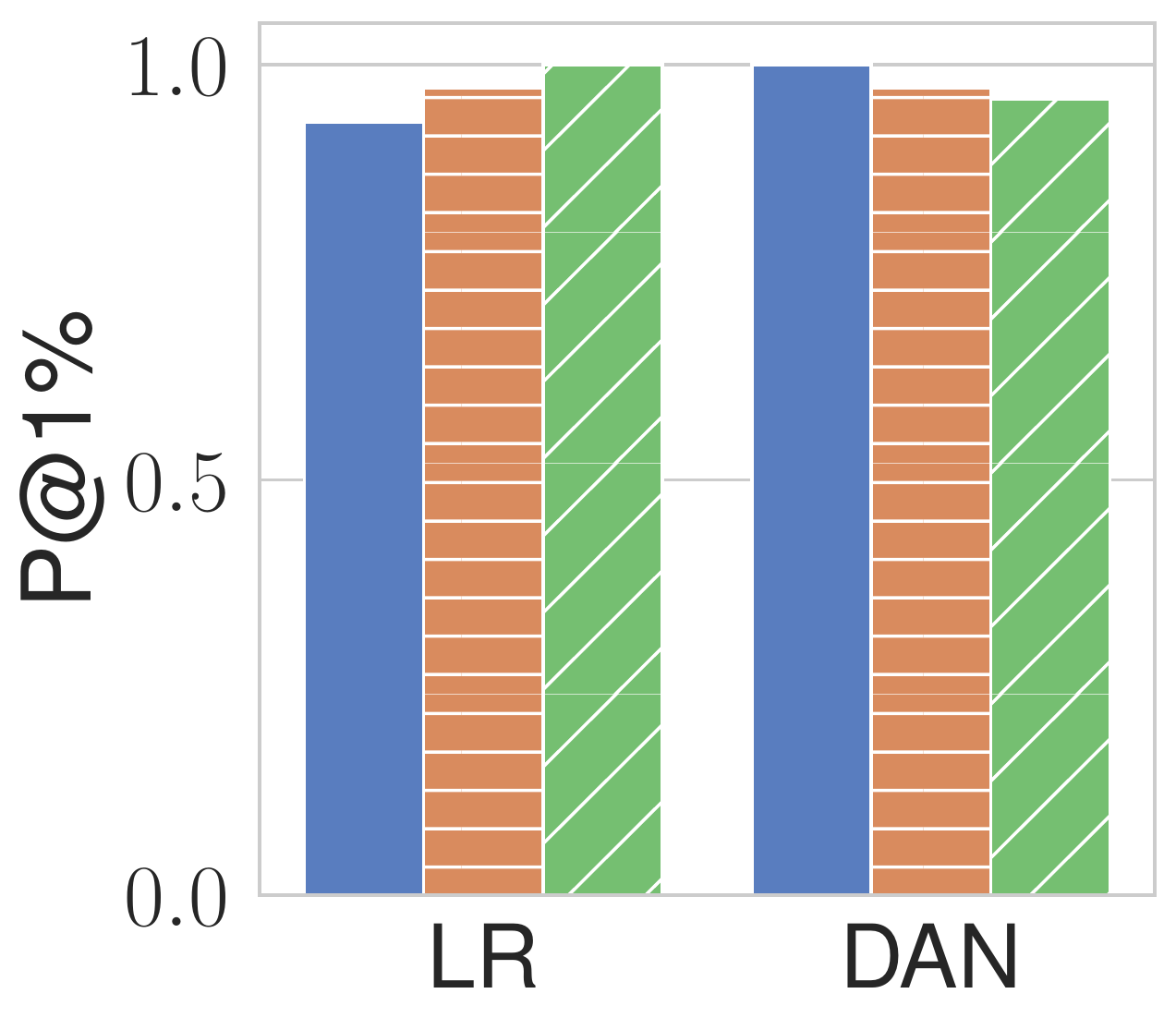}
    \caption{Precision at 1\%.}
    \label{fig:p1-mortality}
\end{subfigure}
\begin{subfigure}[t]{0.24\textwidth}
    \centering
    \includegraphics[trim=0 0 1020 0,clip,width=0.95\textwidth]{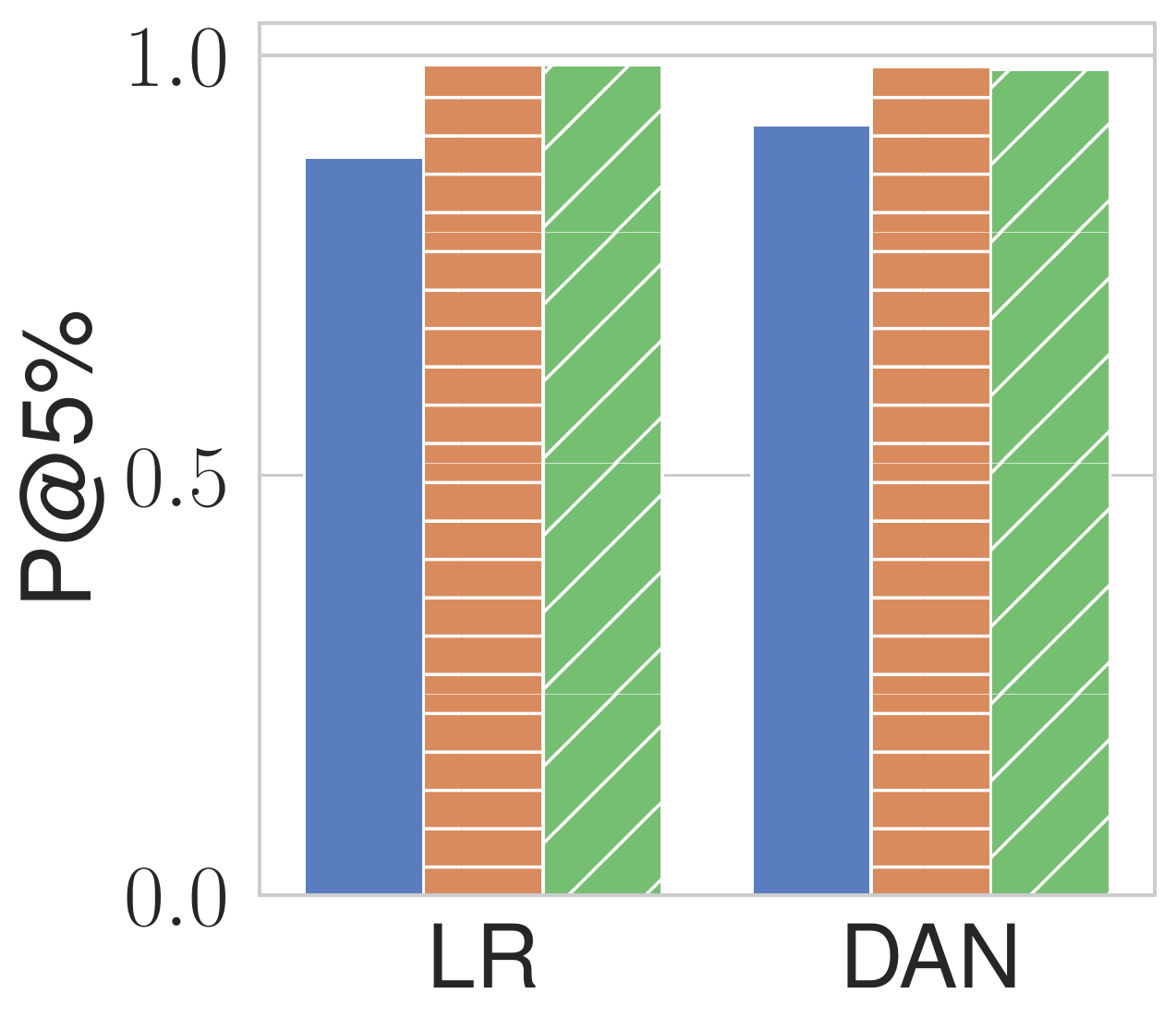}
    \caption{Precision at 5\%.}
    \label{fig:p5-mortality}
\end{subfigure}
  \begin{subfigure}[t]{1\textwidth}
      \centering
      \includegraphics[trim=365 195 0 40,clip,width=0.7\textwidth]{figs/results/mortality-physician-24-PRAUC.pdf}
  \end{subfigure}

  \caption{Results of PR-AUC/ROC-AUC/Precision at 1\%/Precision at 5\% scores on LR/DAN models in mortality prediction (48 hours and retrospective) tasks. 
  Notes are marginally valuable in mortality prediction.
  }
  \label{fig:results}
\end{figure*}

\end{document}